\documentclass{ecai} 

\usepackage{latexsym}
\usepackage{amssymb}
\usepackage{amsmath}
\usepackage{amsthm}
\usepackage{booktabs}
\usepackage{enumitem}
\usepackage{graphicx}
\usepackage{color}
\usepackage{algorithm}
\usepackage{algorithmic}
\usepackage{listings}
\usepackage{graphicx} 
\usepackage{booktabs} 
\usepackage{xcolor}   
\usepackage{multirow}
\usepackage{url}

\newtheorem{definition}{Definition}

\lstdefinestyle{text}{
  basicstyle=\ttfamily\scriptsize,
  breaklines=true,
  breakindent=0pt,
  breakautoindent=false,
  breakatwhitespace=true,
  columns=fullflexible,
  frame=single,
  xleftmargin=0pt,
  xrightmargin=0pt
}

\lstdefinestyle{pythonstyle}{
  language=Python,
  basicstyle=\ttfamily\scriptsize,  
  numbers=none,
  numberstyle=\tiny\color{gray},
  stepnumber=1,
  numbersep=5pt,
  frame=single,
  rulecolor=\color{black},
  tabsize=2,
  captionpos=b,
  breaklines=true,
  breakindent=0pt,
  breakautoindent=false,
  breakatwhitespace=true,
  keywordstyle=\color{blue},
  commentstyle=\color{teal},
  stringstyle=\color{red},
  columns=flexible,  
  xleftmargin=\fill,   
  xrightmargin=\fill   
}

\definecolor{darkgreen}{rgb}{0.0, 0.5, 0.0} 
\definecolor{cornflower}{RGB}{147, 204, 234}
\definecolor{salmon}{RGB}{246, 146, 137}

\newcommand{\BibTeX}{B\kern-.05em{\sc i\kern-.025em b}\kern-.08em\TeX}

\begin{document}


\begin{frontmatter}


\paperid{123} 


\title{ELATE: Evolutionary Language model for Automated Time-series Engineering}


\author{\fnms{Andrew}~\snm{Murray}}
\author{\fnms{Danial}~\snm{Dervovic}} 
\author{\fnms{Michael}~\snm{Cashmore}}

\address{JP Morgan AI Research}


\begin{abstract}
Time-series prediction involves forecasting future values using machine learning models. Feature engineering, whereby existing features are transformed to make new ones, is critical for enhancing model performance, but is often manual and time-intensive. Existing automation attempts rely on exhaustive enumeration, which can be computationally costly and lacks domain-specific insights. We introduce ELATE (Evolutionary Language model for Automated Time-series Engineering), which leverages a language model within an evolutionary framework to automate feature engineering for time-series data. ELATE employs time-series statistical measures and feature importance metrics to guide and prune features, while the language model proposes new, contextually relevant feature transformations. Our experiments demonstrate that ELATE improves forecasting accuracy by an average of 8.4\% across various domains.
\end{abstract}

\end{frontmatter}


\section{Introduction}

Time-series data is composed of a sequence of data-points measured over time. The problem of time-series prediction (often referred to as forecasting) involves using past data to predict how target variables will evolve in the future. Time-series prediction problems are ubiquitous in society: for example predicting stock price movement \cite{roondiwala2017predicting}, forecasting global warming \cite{mudelsee2019trend} or understanding disease progression in patients \cite{nguyen2020predicting}.

Traditionally, time-series prediction problems have been tackled using parametric models defined by a fixed set of statistically derived parameters \cite{box2015time}. Such approaches typically rely on strong assumptions on the underlying distribution of the data. In recent years, machine learning (ML) models have become increasingly popular for time-series prediction tasks due to their ability to learn complex, non-linear patterns in the data \cite{lim2021time}. 

However, implementing new ML models is time consuming and involves many manual steps: data cleaning, pre-processing, feature engineering, hyper-parameter tuning and model evaluation. In feature engineering (FE), raw time-series features are combined and transformed into a set of informative features with stronger predictive capability. Many practitioners note that FE often has a greater impact on model performance than the choice of model itself \cite{verdonck2024special}. On the other hand, it also demands the most substantial time commitment. Amazon estimates that more than 50\% of data scientists time is consumed in FE related tasks \cite{AmazonOps}. Automated feature engineering (Auto-FE) seeks to reduce the time to production of new ML models by automating this process using computational techniques~\cite{chen2021techniques}.

Prior approaches at tackling Auto-FE have predominantly focused on tabular data and typically do one of the following: enumerate all features given a predefined set of transformations and then perform feature selection \cite{christ2018time}; or use optimization algorithms to find the optimal transformations to perform. Enumeration can be computationally costly, in addition the space of possible transformations is infinite and so the generated features must be constrained by limiting the transformations used \cite{de2022tsfuse}. This requires some initial domain understanding, we must know what type of transformations are likely to be useful for the given domain. Optimization approaches face a similar drawback, the search tree grows exponentially and so clever heuristics are required to explore the solution space efficiently. This is exacerbated in time-series forecasting, since we must not only select transformations, but also determine the span of past time periods over which to apply them \cite{fulcher2018feature}. As an example, computing the rolling mean of the past 7 days provides very different information to the rolling mean of the past 30 days.

Recent work has successfully applied large language models (LLM) to the problem of automated feature engineering for tabular data \cite{hollmann2024large,gong2024evolutionary}. Expanding upon this idea, we introduce ELATE, a novel approach to time-series automated feature engineering (TS Auto-FE), which utilizes the extensive domain knowledge of the LLM as a heuristic. This approach uses time-series feature importance measures to evaluate and prune low-scoring features within an evolutionary optimization framework, guiding the search for optimal features. We perform a comprehensive experimental evaluation, comparing against 8 baselines on 7 time series prediction tasks. Our results show that ELATE outperforms all baselines in terms of root mean squared error (RMSE) and mean absolute error (MAE), with average reductions of 8.4\% and 9.6\% compared to no FE. This process completes in hours, a task that would take data scientists days.

In section \ref{sec:encoding}, we define the TS Auto-FE problem.
In section \ref{sec:related} we place the contribution of this paper in context with respect to related work.
In section~\ref{sec:method} we outline the ELATE procedure.
In section~\ref{sec:experiments} we describe the setup and results of our experimental evaluation.
We conclude and address avenues for future research in section~\ref{sec:conclusion}.

\section{Problem Definition}\label{sec:encoding}
A time-series is a sequence of values measured over time: $x = [x_1,x_2,\dots,x_m]$, where $x_i \in \mathbb{R}$ is the value of $x$ at time $t_i$. A multi-variate time-series forecasting problem can be framed as a regression problem in which the input is a dataset containing many time-series features:

\begin{definition}[Multi-variate Time-series Forecasting] A multi-variate time-series forecasting problem can be modeled as a regression problem with input dataset $D = [X, y]$, where $X \in \mathbb{R}^{m\times n}$ is the feature matrix; such that each column $X_{:,j}$ is a time-series feature and each row $X_{i,:}$ is a sample corresponding to a time $t_i$.
Our goal is to learn a function $f$, such that $y \approx f(X)$ and $[f(X)]_t$ can only use inputs $X_{t', :}$, where $t'< t$.
\end{definition}

In TS Auto-FE, the goal is to automatically engineer a new feature matrix $X'$, which has greater predictive performance than $X$. New features can be generated by applying transformations to existing ones. 

\begin{definition}[Feature Transformation]
A feature transformation $T : \mathbb{R}^{m\times n} \rightarrow \mathbb{R}^{m}$, is a function applied to the feature matrix $X$, which generates a new feature $X_{:,n+1}$.
\end{definition}

We use $\mathcal{T} = \{T_1, T_2,\dots,T_p\}$ to refer to a set of transformations of size $p$. We define the transformation composition operator $\phi(\mathcal{T},X) = \left[X, T_1(X), T_2(X),\dots, T_p(X)\right]$ which performs the columnwise concatenation of each transformation $T_k \in \mathcal{T}$ on $X$, to generate a new feature matrix $X' \in \mathbb{R}^{m \times (n + p)}$ with $p$ additional features. Note, that the order of the transformations in $\phi$ does not matter, since each transformation $T_k \in \mathcal{T}$, is a function purely of the base features from $X$. That is to say, the transformation $T_1 = (X_{:,j} + X_{:,k})$ would be treated independently from $T_2 = 2(X_{:,j} + X_{:,k})$.

Finally, if we define $E : (X', y) \rightarrow \mathbb{R}$, as an \textit{evaluation metric} which evaluates the prediction error of a feature matrix $X'$ in predicting the target time-series $y$; then we can frame TS Auto-FE as an optimization problem:

\begin{definition}[Time-series Automated Feature Engineering Problem]
A TS Auto-FE problem is an optimization problem:
$$
\min_{\mathcal{T}} E(\phi(\mathcal{T}, X), y)
$$
\end{definition}
The solution to a TS Auto-FE problem is the set of transformations $\mathcal{T}$ minimizing $E$. Henceforth, we use the notation $X_j = X_{:,j}$ to refer to a specific feature at index $j$.

\section{Related Work}\label{sec:related}
Here we review the relevant literature related to TS Auto-FE. It should be noted that there has been a plethora of recent work using LLMs to make forecasts directly \cite{jin2023time,tan2024language}. In this paper we are primarily interested in solving the problem of \textit{automated feature engineering} for time-series, as opposed to evaluating different methodologies for forecasting. For an overview of such techniques we refer the reader to a relevant survey \cite{zhang2024large}.

\paragraph{Expand and Reduce}
The expand and reduce approach is a popular method in Auto-FE, involving exhaustive feature generation (expansion) followed by pruning (reduction) using feature selection techniques \cite{kanter2015deep,lam2017one,horn2020autofeat}. Several papers address feature extraction for time-series \textit{classification} within an expand and reduce framework \cite{barandas2020tsfel,lubba2019catch22,fulcher2017hctsa,hyndman2019tsfeatures,kaul2017autolearn}. These methods compute summary statistics from entire time-series, mapping each series to a row in the extracted data-frame, which is used to train a classifier. In contrast, Auto-FE for forecasting involves creating new columns for each transformation. \texttt{Vest} \cite{cerqueira2021vest} and \texttt{tsfresh} \cite{christ2018time} present expand and reduce approaches for forecasting, by performing transformations across lagged values. The former is only capable of handling uni-variate time-series, while the latter is both time and memory intensive. Both approaches are also incapable of handling more complex transformations composed of a number of sequential steps.

\paragraph{Evolutionary Approaches}
Evolutionary approaches maintain a set of candidate features which are iteratively improved via mutation and selection \cite{smith2003feature,dor2012strengthening,katz2016explorekit}. These approaches rely on sampling transformations, in contrast ELATE uses the LLM to suggest transformations it believes will be useful for the given forecasting task.

\paragraph{Search Based Methods}
In search based methods, a transition tree is constructed, where each node represents a dataset with specific features, and edges represent transformations leading to new feature sets. Khurana et al. \cite{khurana2016cognito} explore search strategies in \texttt{Cognito}. Reinforcement learning has also been proposed to learn a policy for traversing the transformation tree \cite{khurana2018feature,li2022learning}. The main drawback of these approaches is the massive branching factor leading to a combinatorial explosion of states.

\paragraph{Neural Network Methods}
A number of authors have investigated using neural networks for Auto-FE \cite{nargesian2017learning,chen2019neural,zhu2022difer}. These approaches learn a neural network based on the supplied data, meaning that the learned model is not transferable to new tasks. In contrast, ELATE is generalizable. Deep learning approaches automatically learn new features within layers of the neural network, however these learned features are not interpretable and are therefore unsuitable for many applications. On the other hand, ELATE explicitly returns the code required to generate the feature, as well as a description of its potential utility.

\paragraph{Large Language Model Approaches}
All of these prior approaches are missing one key ingredient that make human data scientists so effective at FE: they lack \textit{context} regarding the prediction task being performed. Transformation sequences that work well for one domain are not guaranteed to work well for others. This makes \textit{learned} approaches not generaliseable, whereas \textit{generic} search strategies are too inefficient without clever heuristics to guide them. Recently, LLMs have been proposed as an effective tool in Auto-FE due to their ability to understand the wider context of the task. LLMs are neural networks that have been trained on vast amounts of data and have been applied successfully to a broad spectrum of tasks: from question-answering to writing code \cite{bubeck2023sparks}. Hollman et al. \cite{hollmann2024large} leverage LLMs to automatically generate features for tabular datasets. An interpreter is used in-the-loop to ensure that the code output by the LLM is valid. Gong et al. \cite{gong2024evolutionary} embed an LLM within an evolutionary framework for tabular data. 

These past works differ from ELATE in a number of ways. The work of Hollman et al. uses a simple iterative procedure for feature generation, where new features are kept only if they improve performance. This approach ignores optimization trajectory and may prematurely discard features that could be beneficial in combination with future ones. Gong et al. limit transformations to a fixed set of primitive operators applied sequentially, which may overlook valuable transformations that don't show immediate benefit. As an example, in a model predicting whether a patient will develop diabetes, the body mass index (BMI) might be a useful feature and can be found by the following sequence of transformations, $\{height^2 = height \times height, BMI = weight / height^2\}$. The optimization may explore the transformation $height^{2} = height \times height$, not observe any reward, thus preventing the procedure from generating the feature BMI. In contrast, ELATE can understand the context (BMI is an indicator of how healthy someone is, which can be useful for determining if someone will develop diabetes) without having to reason over the intermediate steps. Finally, these approaches are designed for tabular data. Auto-FE for tabular data involves transforming and combining static features without the need to account for temporal order or time-based relationships. In contrast, TS Auto-FE involves capturing temporal dependencies and patterns, such as trends, seasonality, and lagged values~\cite{fulcher2018feature}. Aggregations such as rolling means are much more difficult to model using primitive numeric operations. As far as we are aware, ELATE is the first method for TS Auto-FE that leverages LLMs.

Finally, these approaches are designed for tabular data. Auto-FE for tabular data involves transforming and combining static features without the need to account for temporal order or time-based relationships~\cite{galli2024python}. In contrast, TS Auto-FE involves capturing temporal dependencies and patterns, such as trends, seasonality, and lagged values~\cite{fulcher2018feature}. Aggregations such as rolling means are much more difficult to model using primitive numeric operations. As far as we are aware, ELATE is the first method for TS Auto-FE using LLMs.

\section{Method}\label{sec:method}
\begin{figure}
    \centering
    \includegraphics[width=\columnwidth]{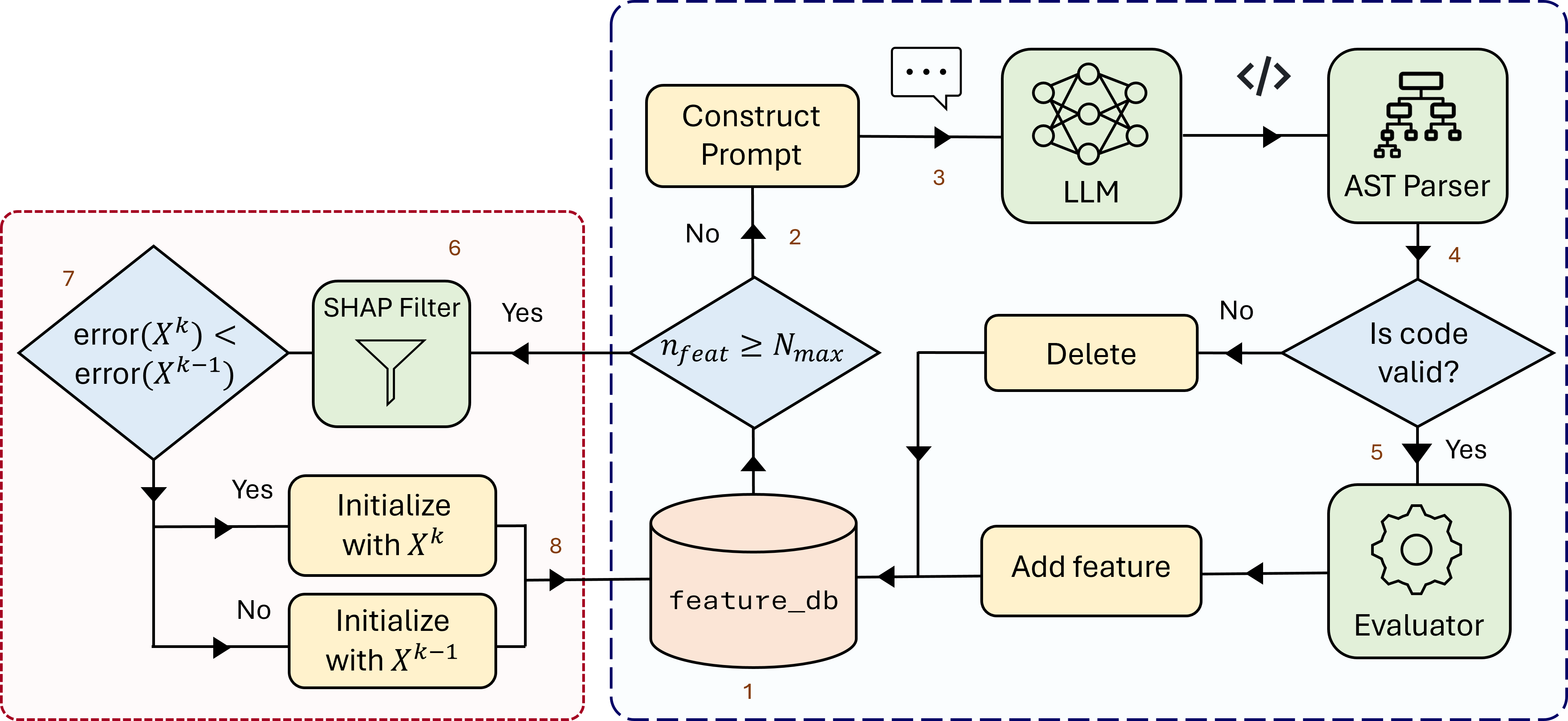}
    \caption{Figure showing architecture of ELATE. Features are stored in the \texttt{feature\_db}. The \textcolor{blue}{right} loop (1-5) adds new members to the population, while the \textcolor{red}{left} loop (6-8) prunes low scoring members at the end of each generation.}
    \label{fig:llmfs}
\end{figure}

The ELATE architecture is shown in Figure \ref{fig:llmfs} and is inspired by Romera-Paredes et al. \cite{romera2024mathematical}. The input to the procedure is the following:  the \textit{prompt template} summarizing the task to be performed by the LLM, the \textit{dataset description} containing metadata describing the data (features names, descriptions, data types and whether they contain NaNs), a set of \textit{initial features} $X_I = \{X_1,\dots,X_{n_{prompt}}\}$ and the \textit{base dataset} $D_B = \left[X_B, y\right]$, containing the base features $X_B$ and the target $y$. The parameters are as follows: the maximum number of features in the feature database (\texttt{feature\_db}) $N_{max}$, the size of the best feature set $N$ (this is the size of the \texttt{feature\_db} after pruning), the number of generations $G$ and the number of example features to include in the prompt $n_{prompt}$.

The process works by maintaining a population of features within the \texttt{feature\_db}. The \texttt{feature\_db} is initialized using $X_I$. In our experiments, we used two LLM generated features. On each generation $k$, we store the $N$ best features, $X^k \in \mathbb{R}^{m\times N}$.

Referring to Figure \ref{fig:llmfs}, in (1), the \texttt{feature\_db} is queried and $n_{prompt}$ features are sampled. Features with higher evaluator scores are assigned higher probabilities in the sampling procedure. The sampled features are combined with the prompt template and dataset description to construct a prompt asking the LLM to generate a new feature in the form of Python code (2). This is then passed to a LLM (3), and the responses stored. Responses are parsed for code validity using abstract syntax tree (AST) parsing (4). If the code is found to be valid, it is compiled and executed resulting in a feature series $X_j$, which is passed to the evaluator (5). The evaluator returns a value, which is then stored alongside the feature in the \texttt{feature\_db}, returning to (1). If the code is invalid, the evaluator is bypassed and we return directly to (1). 

This process repeats (1 - 5), until the number of features stored in the \texttt{feature\_db} reaches $N_{max}$. At this point, a round of selection is undertaken using a SHapley Additive ex-Planations (SHAP) feature filter \cite{SHAP} (6). The top $N$ features with the highest magnitude SHAP value are selected as the best feature set for the current generation. These are not guaranteed to result in the best $N$ features in terms of predictive performance. As a result, we perform an additional check in (7) to see if the RMSE of the new best feature set $X^k$, is better than the previous best feature set $X^{k-1}$. If so, we use $X^k$ to initialise the \texttt{feature\_db} in the next generation (8), otherwise, we use $X^{k-1}$. This process repeats for $G$ generations. A more detailed overview of the relevant components is provided below.

\paragraph{Prompt Construction}
The prompt template used in the experiments is provided in Figure \ref{fig:prompt_template}. The placeholder \texttt{@@description@@} is replaced with the dataset description containing the relevant metadata. This provides the LLM with context relevant to the specific prediction task (for example feature names, descriptions and data types). Example dataset descriptions are provided in Appendix \ref{app:description} for each domain.

\begin{figure}[t]
    \centering
    \resizebox{0.8\columnwidth}{!}{%
        \begin{minipage}{\linewidth}
            \lstset{style=text}
\begin{lstlisting}
You are a data scientist tasked with developing a time series prediction model.
I will now provide you with a description of the dataset along with the input and target features names and data types.

@@description@@

You must use only the INPUT FEATURES to engineer new features by performing transformations, you cannot use the TARGET column.
I want you to provide me with a block of Python code that takes the dataset as input as a Pandas dataframe and makes a variable called feature with the new feature values.
You should add as a comment before the code a description of the feature and its potential utility to the prediction task being performed.

The code should be in the following format:

# Description
# Here you should add a comment outlining what the feature represents and why it might be useful.
df['new_feature'] = Your code here
feature = df['new_feature']

- You must only use the input features to generate new features (you cannot use the target).
- Do not use negative shifts on existing features, e.g. df[existing_feature].shift(-1).
- Where possible you should fill any NaN with an appropriate value, but avoid using bfill.
- When performing aggregations over past values in a column, you must only use previous rows data. One way this can be achieved is by using rolling functions with a specified window size that includes only past data, such as rolling().mean(), rolling().sum() as opposed to .mean() or .sum().
- You must ensure that the returned Series has the same index as the original DataFrame.
- You are permitted to use the packages Numpy, Sklearn, Math, Pandas, Scipy.stats, and Statsmodels.tsa along with core Python functions.
- You must only provide the Python code for one feature, without any additional text, code tags, code span, markdown formatting or backticks as I need to be able to execute your response.
- You must engineer features that have not been generated previously.

I will now provide you with some examples. You should try to come up with one that is better than the examples.

@@examples@@

You can use and combine ideas from previously generated features to engineer new ones. Each previously generated feature is evaluated and their score can be used to guide you towards useful transformations. Higher scores are better. Features and scores are given below in the form (FEATURE_NAME, SCORE).

@@generated_features@@
\end{lstlisting}
\begin{lstlisting}
This dataset can be used to predict transformers' oil temperature. This helps manage electric power distribution efficiently and prevent waste and equipment depreciation. The input features include various power load metrics, with the oil temperature as the target for prediction.

The input features and target to be predicted are:

# INPUT FEATURES

HUFL (float64): High UseFul Load.
HULL (float64): High UseLess Load.
MUFL (float64): Middle UseFul Load.
MULL (float64): Middle UseLess Load.
LUFL (float64): Low UseFul Load.
LULL (float64): Low UseLess Load.
Year (int32): Year.
Month (int32): Month.
Day (int32): Day.
Hour (int32): Hour.
OT_Tminus1 (float64): Current oil temperature indicating the condition of the electrical transformer.

# TARGET

OT (float64): Oil Temperature one hour in advance.
\end{lstlisting}
        \end{minipage}
    }
    \caption{Prompt template used within ELATE (top) and an example dataset description for the ETTh1 domain (bottom).}
    \label{fig:prompt_template}
\end{figure}

We use few-shot prompting \cite{brown2020language} by providing example feature transformations sampled from the \texttt{feature\_db}. This helps the LLM to understand the structure of the output code. The placeholder \texttt{@@examples@@} is replaced by the features sampled from the \texttt{feature\_db}.

We also update a section called \texttt{@@generated features@@} within the prompt template, each time a new feature is generated. The names and evaluator scores of the most recent 250 features generated are added here. This offers a number of benefits: first, it prevents the LLM generating features that it has already generated, second it guides the trajectory of the optimization by helping the LLM identify which prior transformations have been effective.

\paragraph{Evaluator}
The evaluator $E$, is a function $E : (X_j, Y) \rightarrow \mathbb{R}$, which provides ELATE with an estimate of the predictive power of the new time-series feature on the target. Since this is computed each time the \texttt{feature\_db} is sampled, it must be computationally efficient. In addition, we require a feature importance measure which does not change with the addition of new features. This excludes feature selection techniques such as SHAP which are influenced by the presence of additional features (the SHAP value of a feature $X_j$ can change from one iteration to another as we add new features to the \texttt{feature\_db}).

As a result, we chose to use two different time-series statistical measures, each capturing unique interactions between the feature and target: Granger causality $E_{GC}$ and mutual information $E_{MI}$. Granger Causality \cite{shojaie2022granger} captures temporal and causal linear interactions between two time-series. To account for non-linear interactions, we use mutual information. The overall evaluator score is the mean of the two evaluators. A detailed overview of the individual evaluators is provided in Appendix \ref{app:eval}.

\paragraph{Feature Sampling}
As per Romera-Paredes et al. \cite{romera2024mathematical}, probabilities are assigned to each feature based on their evaluator scores. Unlike Romera-Paredes et al., we chose to use an exponentially decaying function to model the temperature $T$, at any iteration, as a function of the number of features $n_{feat}$, currently in the \texttt{feature\_db}: $T = T_0 e^{\frac{-K n_{feat}}{N_{max}}} + \epsilon$, where $T_0$, $K$ and $\epsilon$ are constant parameters. The temperature exponentially decays over the course of a generation, while the constant $K$ can be used to vary the rate of decay. We use $T_0 = 10$, $K = 5$ and $\epsilon = 0.1$ in our experiments.

Given the temperature $T$ at any given iteration, we can compute the probability of sampling a feature $X_j$ as: $P_{X_j} = e^{\left(E(X_j, Y) / T\right)}/\sum_{X_k \in X^{F}}e^{\left(E(X_k, Y) / T\right)}$, where $X^F$ is the set of features currently in the \texttt{feature\_db}. The lower the temperature, the higher probability is assigned to features with high evaluator scores. Since the temperature is exponentially decaying, this has the effect of encouraging exploration early on in the generation, and exploitation towards the end.

\paragraph{AST Parser}
The LLM generates Python code as a string, which we validate using an abstract syntax tree (AST) parser. This ensure it adheres to a predefined set of allowable nodes, packages, functions, and attributes, serving as a security measure. Once validated, the code is executed to produce a Pandas \cite{mckinney2011pandas} series, stored with the code string as a feature object. This allows for easy reuse of feature transformations on different datasets by compiling and executing the code with a new dataframe.

\paragraph{SHAP Filter}\label{sec:filter}
When the number of features in the \texttt{feature\_db} reaches the threshold $N_{max}$, feature selection is performed to prune low scoring features. In this case, we want to evaluate the performance of all generated features with respect to all other generated features. Evaluating all subsets of features in a \emph{Remove and Retrain}~\cite{ROAR} manner is computationally infeasible, so we use model feature importance measures as a proxy.
Among feature importance measures, SHAP~\cite{SHAP} has gained popularity due to its theoretical guarantees and efficient implementations for different classes of model. We chose TreeSHAP~\cite{lundberg2018consistent} with interventional SHAP values, as they correspond to ``true-to-the-model'' assessments of feature importance~\cite{chen2020truemodeltruedata}. We use recursive feature elimination to iteratively remove irrelevant features, thereby reducing redundancy and noise. Using XGBoost with walk-forward cross-validation, we compute and normalize SHAP values for each fold and instance. We then compute the mean across all instances and folds. Before pruning, features with a correlation coefficient greater than 0.9 are identified and one of them removed. We then prune the worst 10\% of features based on their SHAP scores, repeating the process until the number of features in the \texttt{feature\_db} equals $N$.

\section{Experiments}\label{sec:experiments}
Our experiments involved 7 different time-series prediction tasks. This enabled us to assess ELATE's versatility and contextual awareness. We prioritized datasets with well-defined feature names and descriptions. Unfortunately, this excluded some forecasting benchmarks, in which feature names and descriptions are not provided \cite{makridakis2020m4}. This context is necessary for the LLM to discern the potential utility of each feature and what it represents. Finally, we tried to identify datasets with different numbers of instances and prediction period horizons. The resulting datasets are summarized in Table \ref{tab:datasets}. Five of the datasets can be found on Kaggle (Store, Energy, Pollution, Trading and Food) and the other two (ETT \cite{zhou2021informer} and ILI \cite{wu2021autoformer}) are widely used forecasting benchmarks.
\begin{table}
    \centering
    \resizebox{\columnwidth}{!}{%
    \begin{tabular}{c c c c c}
        \toprule
        \textbf{Name} & \textbf{Description} & \textbf{\#Instances} & \textbf{\#Features} & \textbf{Horizon} \\
        \midrule
        ILI\footnotemark[1] & Predict new influenza cases & 965 & 10 & 1 Month \\
        Store\footnotemark[2] & Predict store sales & 6300 & 11 & 1 Week \\
        ETTh1\footnotemark[3] & Predict electricity transformer temperature & 17419 & 11 & 1 Hour \\
        Energy\footnotemark[4] & Predict energy demand & 35040 & 21 & 1 Day \\
        Pollution\footnotemark[5] & Predict PM2.5 concentration & 43776 & 13 & 1 Day \\
        Trading\footnotemark[6] & Predict returns of stocks & 125750 & 12 & 1 Day \\
        Food\footnotemark[7] & Predict meal delivery orders & 179568 & 10 & 1 Week \\
        \bottomrule
    \end{tabular}
    }
    \caption{Summary of datasets with their characteristics.}
    \label{tab:datasets}
\end{table}

\footnotetext[1]{\scriptsize https://gis.cdc.gov/grasp/fluview/fluportaldashboard.html}
\footnotetext[2]{\scriptsize https://www.kaggle.com/datasets/yasserh/walmart-dataset/data}
\footnotetext[3]{\scriptsize https://github.com/zhouhaoyi/ETDataset}
\footnotetext[4]{\scriptsize \raggedright https://www.kaggle.com/datasets/nicholasjhana/energy-consumption-generation-prices-and-weather/data}
\footnotetext[5]{\scriptsize \raggedright https://www.kaggle.com/datasets/rupakroy/lstm-datasets-multivariate-univariate/data}
\footnotetext[6]{\scriptsize https://www.kaggle.com/datasets/andrewmvd/sp-500-stocks/data}
\footnotetext[7]{\scriptsize http://www.kaggle.com/datasets/kannanaikkal/food-demand-forecasting/data}

XGBoost was used as our model of choice \cite{chen2016xgboost} with the same model hyper parameters used for each dataset. We use walk-forward cross validation, with the number of folds for each dataset selected in accordance with the size of the test set. Walk-forward validation is a technique used to evaluate the performance of time-series prediction models. We start with an initial portion of the data which is used to train the model. Predictions are then made on a test set immediately following the training data. After evaluating the models performance, the test set is added to the training set, and the test set is updated to use the subsequent portion of data. This mimics the way models are used in real-world scenarios, where predictions are made sequentially over time. For each dataset, we reserve the final 10\% of the dates for testing and the penultimate 10\% for validation (for use within ELATE).

To enable features to be generated using lagged values of the target, we provide ELATE with the column \textit{Target\_Tminus1}. This represents the most recent value of the target variable that would be available at the forecast time. Unlike prior TS Auto-FE methods, which require the user to specify a set of lags over which to generate new features, ELATE has the ability to generate any past lags by performing transformations or aggregations on this column. We run ELATE for 10 generations with $N_{max} = 100$ and $N = 50$. A number of baselines are used for comparison.

\begin{itemize}
    \item T$-$1: A naive prediction model which assumes that there is no change versus the previous time.
    \item Base: Base features (before any FE).
    \item Zero-Shot: Base features plus LLM engineered features without ELATE. These features are generated by passing a zero-shot prompt to GPT-4o.
    \item VEST: Base features plus additional features generated using the python package \texttt{vest} \cite{cerqueira2021vest}.
    \item TSFRESH: Base features plus additional features generated using the rolling time-series mechanism from the python package \texttt{tsfresh} \cite{christ2018time}.
    \item LSTM: Long Short Term Memory (LSTM) neural network trained using the base features. Constant hyperparameter were used throughout: LSTM units = 50, batch size = 32, epochs = 50, timesteps = 10.
    \item ELATE 4o+FRESH: Base features plus additional features generated using ELATE. GPT-4o used as the LLM. FeatuRe Extraction based on basis of Scalable Hypothesis tests (FRESH) algorithm \cite{christ2016distributed} from \texttt{tsfresh} as the feature filter.
    \item ELATE 4o+Const: Base features plus additional features generated using ELATE. GPT-4o used as the LLM. Ablation study in which all features recieve a constant evaluator score.
    \item ELATE 3.5T+SHAP: Base features plus additional features generated using ELATE. GPT-3.5 Turbo used as the LLM. Evaluator and feature filter as per Section \ref{sec:method}.
    \item ELATE 4o+SHAP: Base features plus additional features generated using ELATE. GPT-4o used as the LLM. Evaluator and feature filter as per Section \ref{sec:method}.
\end{itemize}

We use the base features plus the top 50 features for all baselines with the exception of LSTM and T$-$1. Default parameters were used for all other TS Auto-FE packages.

\paragraph{Accuracy}
A comparison of prediction performance in terms of MAE and RMSE is provided in Table \ref{tab:comparison}. In all domains except trading (where it is on par), ELATE 4o+SHAP performs better than T$-$1, Base, Zero-Shot, \texttt{vest}, \texttt{tsfresh} and LSTM in terms of both RMSE and MAE. 

\begin{table*}[t]
    \centering
    \resizebox{.9\textwidth}{!}{%
    \begin{tabular}{l c c c c c c c c}
        \toprule
        \textbf{Method} & \textbf{Metric} & \textbf{ILI} & \textbf{Store} & \textbf{ETTh1} & \textbf{Energy} & \textbf{Pollution} & \textbf{Trading} & \textbf{Food} \\
        \midrule
        \multirow[t]{4}{*}{\centering T$-$1}
        & MAE & 0.0317 $\pm$ 0.0115 & 0.0130 $\pm$ 0.0019 & 0.0088 $\pm$ 0.0004 & 0.1066 $\pm$ 0.0073 & 0.0671 $\pm$ 0.0047 & 0.0097 $\pm$ 0.0003 & 0.0059 $\pm$ 0.0007\\
        & RMSE & 0.0563 $\pm$ 0.0138 & 0.0196 $\pm$ 0.0027 & 0.0132 $\pm$ 0.0006 & 0.1539 $\pm$ 0.0084 & 0.0996 $\pm$ 0.0056 & 0.0142 $\pm$ 0.0004 & 0.0143 $\pm$ 0.0026 \\
        \midrule
        \multirow[t]{4}{*}{\centering Base}
        & MAE & 0.0255 $\pm$ 0.0077 & 0.0113 $\pm$ 0.0016 &  0.0092 $\pm$ 0.0004 & 0.0871 $\pm$ 0.0044 & 0.0604 $\pm$ 0.0046 & 0.0067 $\pm$ 0.0002 & 0.0061 $\pm$ 0.0008 \\
        & RMSE & 0.0427 $\pm$ 0.0097 & 0.0168 $\pm$ 0.0021 &  0.0134 $\pm$ 0.0006 & 0.1145 $\pm$ 0.0050 & 0.0872 $\pm$ 0.0051 & \textbf{0.0100 $\pm$ 0.0003} & 0.0127 $\pm$ 0.0022 \\
        \midrule
        \multirow[t]{4}{*}{\centering Zero-Shot}
        & MAE & 0.0319 $\pm$ 0.0129 & 0.0122 $\pm$ 0.0016 & 0.0087 $\pm$ 0.0004 & 0.0819 $\pm$ 0.0041 & 0.0598 $\pm$ 0.0041 & \textbf{0.0066 $\pm$ 0.0002} & 0.0061 $\pm$ 0.0007 \\
        & RMSE & 0.0621 $\pm$ 0.0159 & 0.0174 $\pm$ 0.0020 & 0.0129 $\pm$ 0.0006 & 0.1082 $\pm$ 0.0046 & 0.0837 $\pm$ 0.0046 & \textbf{0.0100 $\pm$ 0.0003} & 0.0126 $\pm$ 0.0021\\
        \midrule
        \multirow[t]{4}{*}{\centering VEST}
        & MAE &  0.0311 $\pm$ 0.0118 & 0.0105 $\pm$ 0.0013 & 0.0085 $\pm$ 0.0004 & 0.0663 $\pm$ 0.0045 & 0.0593 $\pm$ 0.0041 & \textbf{0.0066 $\pm$ 0.0002} & 0.0058 $\pm$ 0.0007\\
        & RMSE & 0.0559 $\pm$ 0.0137 & 0.0153 $\pm$ 0.0017 & 0.0127 $\pm$ 0.0006 & 0.0960 $\pm$ 0.0052 & 0.0837 $\pm$ 0.0046 & \textbf{0.0100 $\pm$ 0.0003} & 0.0131 $\pm$ 0.0023 \\
        & Runtime (s) & 220 & 1509 & 3738 & 6985 & 10065 & 27869 & 45912\\
        \midrule
        \multirow[t]{4}{*}{\centering TSFRESH}
        & MAE & - & 0.0121 $\pm$ 0.0012 & - & - & - & - & - \\
        & RMSE & - & 0.0168 $\pm$ 0.0016 & - & - & - & - & -\\
        & Runtime (s) & - & 1791 & - & - & - & - & -\\
        \midrule
        \multirow[t]{4}{*}{\centering LSTM}
        & MAE & 0.0562 $\pm$ 0.0132 & 0.0170 $\pm$ 0.0018 & 0.0117 $\pm$ 0.0006 & 0.0874 $\pm$ 0.0047 & 0.0647 $\pm$ 0.0046 & 0.0067 $\pm$ 0.0002 & - \\
        & RMSE & 0.0788 $\pm$ 0.0866 & 0.0222 $\pm$ 0.0729 & 0.0171 $\pm$ 0.0046 & 0.1205 $\pm$ 0.0259 & 0.0928 $\pm$ 1.7663 & 0.0101 $\pm$ 0.0005 & - \\
        & Runtime (s) & \textbf{69} & \textbf{308} & 8070 & 16017 & 20142 & 57205 & - \\
        \midrule
        \multirow[t]{4}{*}{\centering ELATE 4o+FRESH}
        & MAE & 0.0273 $\pm$ 0.0122 & \textbf{0.0101 $\pm$ 0.0013} & 0.0082 $\pm$ 0.0004 & 0.0747 $\pm$ 0.0039 & 0.0602 $\pm$ 0.0040 & \textbf{0.0066 $\pm$ 0.0002} & 0.0058 $\pm$ 0.0007 \\
        & RMSE & 0.0521 $\pm$ 0.0136 & 0.0149 $\pm$ 0.0017 & 0.0124 $\pm$ 0.0006 & 0.0999 $\pm$ 0.0044 & 0.0842 $\pm$ 0.0045 & \textbf{0.0100 $\pm$ 0.0003} & 0.0124 $\pm$ 0.0022\\
        & Runtime (s) & 1250 & 1280 & \textbf{2721} & \textbf{1824} & \textbf{1711} & \textbf{9899} & \textbf{3466}\\
        & Cost (\$) & 8.60 & 7.83 & 6.92 & 6.00 & 6.21 & 7.49 & 6.54\\ 
        \midrule
        \multirow[t]{4}{*}{\centering ELATE 4o+Const}
        & MAE & 0.0246 $\pm$ 0.0082 & 0.0109 $\pm$ 0.0010 & 0.0083 $\pm$ 0.0004 & 0.0800 $\pm$ 0.0041 & 0.0611 $\pm$ 0.0041 & \textbf{0.0066 $\pm$ 0.0002} & 0.0059 $\pm$ 0.0008\\
        & RMSE &  0.0423 $\pm$ 0.0096 & 0.0156 $\pm$ 0.0015 & \textbf{0.0123 $\pm$ 0.0006} & 0.1059 $\pm$ 0.0046 & 0.0841 $\pm$ 0.0045 & \textbf{0.0100 $\pm$ 0.0003} & 0.0128 $\pm$ 0.0022 \\
        & Runtime (s) & 1488 & 2076 & 5919 & 6898 & 3489 & 12816 & 18216\\
        & Cost (\$) & 9.82 & 6.44 & 7.43 & 6.35 & 6.57 & 6.07 & 6.49\\ 
        \midrule
        \multirow[t]{4}{*}{\centering ELATE 3.5T+SHAP}
        & MAE & 0.0232 $\pm$ 0.0078 & 0.0108 $\pm$ 0.0013 & 0.0086 $\pm$ 0.0004 & 0.0702 $\pm$ 0.0042 & 0.0603 $\pm$ 0.0042 & \textbf{0.0066 $\pm$ 0.0002} & 0.0058 $\pm$ 0.0006 \\
        & RMSE & 0.0397 $\pm$ 0.0094 & 0.0156 $\pm$ 0.0018 & 0.0128 $\pm$ 0.0006 & 0.0970 $\pm$ 0.0047 & 0.0835 $\pm$ 0.0046 & \textbf{0.0100 $\pm$ 0.0003} & 0.0124 $\pm$ 0.0021 \\
        & Runtime (s) & 1419 & 1506 & 3703 & 6749 & 10015 & 12315 & 38460\\
        & Cost (\$) & \textbf{2.07} & \textbf{1.58} & \textbf{1.18} & \textbf{1.27} & \textbf{1.28} & \textbf{1.31} & \textbf{1.68}\\ 
        \midrule
        \multirow[t]{4}{*}{\centering ELATE 4o+SHAP}
        & MAE &\textbf{0.0222 $\pm$ 0.0060} & 0.0104 $\pm$ 0.0010 & \textbf{0.0081 $\pm$ 0.0004} & \textbf{0.0660 $\pm$ 0.0043} & \textbf{0.0592 $\pm$ 0.0040} & \textbf{0.0066 $\pm$ 0.0002}  & \textbf{0.0057 $\pm$ 0.0006}\\
        & RMSE & \textbf{0.0386 $\pm$ 0.0084} & \textbf{0.0148 $\pm$ 0.0015} & \textbf{0.0123 $\pm$ 0.0006} & \textbf{0.0933 $\pm$ 0.0049} & \textbf{0.0820 $\pm$ 0.0044} & \textbf{0.0100 $\pm$ 0.0003} & \textbf{0.0121 $\pm$ 0.0020}\\
        & Runtime (s) & 2003 & 2945 & 4134 & 6342 & 8238 & 19766 & 34677\\
        & Cost (\$) & 13.43 & 8.27 & 7.10 & 5.99 & 6.51 & 7.51 & 6.83\\ 
        \bottomrule
    \end{tabular}%
    }
    \caption{Baseline Comparison (\textbf{Bold is Best}) (- Memory limit exceeded)}
    \label{tab:comparison}
\end{table*}

We assigned 128GB of RAM to run \texttt{tsfresh} and this was exceeded in 6/7 of the domains. This reinforces a key issue with expand and reduce approaches (see Section \ref{sec:related}). As opposed to ELATE which maintains a small set of high quality features, expand and reduce approaches exhaustively apply every transformation to every possible feature subset. This results in excessive memory consumption. Even on the store dataset, which has 11 base features and 6300 rows, running \texttt{tsfresh} resulted in 7831 features with a peak memory consumption of $~13.7$GB. Since \texttt{vest} is tailored to uni-variate prediction tasks, it is only generating features for lagged values of the target, hence it uses less memory than \texttt{tsresh}.

\begin{figure*}
    \centering
    \includegraphics[width=.8\textwidth]{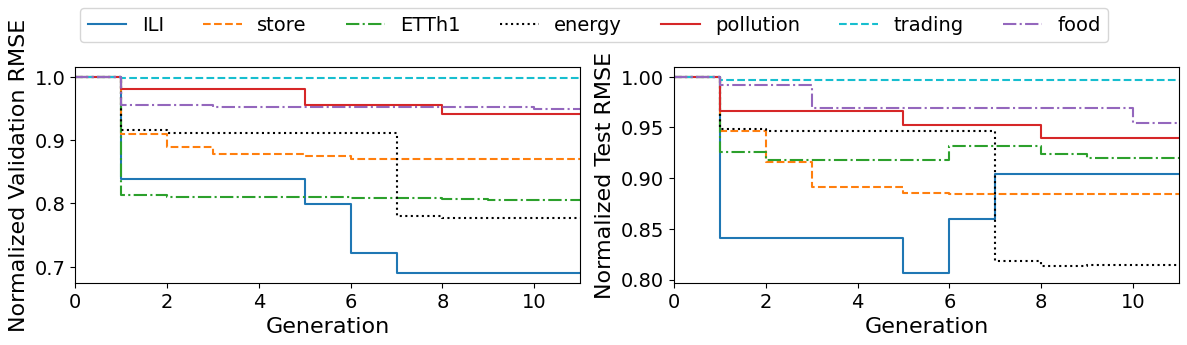}
    \caption{Figure showing normalized RMSE for both the validation (left) and test (right) sets across each domain, plotted against the generation number. RMSE values are normalized relative to their values prior to feature engineering (generation 0)}
    \label{fig:time_features}
\end{figure*}

Figure \ref{fig:time_features} shows the normalized RMSE for both the validation and test sets across each domain, plotted against the generation number. The RMSE values are normalized relative to their values prior to feature engineering (generation 0). In some problems (ETTh1 and ILI), we observe that ELATE is overfitting to the validation set. A number of changes could be made to combat this, examples include early stopping or creating a validation set which is more representative of the test data. We would like to stress, that our goal was not to determine the most predictive model and set of hyper-parameters for each domain; rather it was to perform a fair comparison using the same data pre-processing steps and a uniform set of parameters for all domains, therefore such mitigation steps were not taken.

The improvement in prediction performance varies substantially, from 18.5\% reduction in RMSE versus Base in the energy domain, to no improvement in the trading domain. This is thought to be partially attributed to problem dynamics: stock returns are highly volatile, as opposed to energy consumption which exhibits stronger seasonality and is thus easier to predict (less energy is expected to be used during the day when people are working for example). Nonetheless, we see an average reduction of 9.6\% and 8.4\% in MAE and RMSE respectively. To get a sense of the impact of this, a survey of the consumer goods industry found that a 10\% improvement in forecast accuracy can correspond to a 5\% reduction in inventory costs and a 2\% increase in revenue \cite{mckinsey2018ai}.

\paragraph{Cost} Table \ref{tab:comparison} also compares ELATE's performance using two LLMs: GPT-3.5 Turbo (a cost-effective model) and GPT-4o (a state-of-the-art model). 
GPT-4o resulted in a greater reduction in MAE and RMSE in all domains except trading, with an average reduction of 9.6\% and 8.4\% respectively, versus 5.8\% and 6.6\% for GPT-3.5 Turbo. Nonetheless, using GPT-3.5 Turbo still resulted in a significant average improvement versus the base features at a fraction of the cost  (\$0.50/\$1.5 per million input/output tokens for GPT-3.5 Turbo versus \$2.50/\$10 per million for GPT-4o) \cite{openai_2025}. It is important to point out, that this reduction in cost is somewhat impacted by the proportion of failed prompts. GPT-3.5 Turbo generated more invalid features. More invalid features means that more prompts are necessary, which in turn leads to higher costs. However, the average cost of GPT-3.5 Turbo was still significantly less than GPT4o (\$1.48 versus \$7.95). It is also worth mentioning here that the cost of having a data scientist perform this task is likely to be an order of magnitude more (around \$75/hr \cite{365datascience}).

\paragraph{Runtime}
A comparison of the runtime of ELATE versus the baselines is also provided in Table \ref{tab:comparison}. ELATE 4o+SHAP takes less time than VEST and LSTM on 4/7 domains with an average runtime of approximately 3 hours. In some instances, training a neural network such as LSTM could be faster than running ELATE, however this comes at the expense of interpretability. Furthermore, ELATE is an anytime algorithm. In theory, the user can choose to run ELATE for as many generations as needed, and return the best feature set found within that time. Referring to Figure \ref{fig:time_features}, even after a few generations, ELATE achieves significant reduction in RMSE versus the base features. It is also worth mentioning that it would take a data scientist significantly longer to design all of the features manually \cite{AmazonOps}. 

A large proportion of the runtime of ELATE is spent in the SHAP filter. ELATE is required to compute evaluator scores each time a feature is added, \textit{in addition} to computing SHAP values. To counteract this, we also propose an alternative configuration (ELATE 4o+FRESH), which uses the FRESH algorithm from \texttt{tsfresh} as the feature filter. In this approach, statistical significance tests \cite{sen1968estimates,massey1951kolmogorov} are used in place of the evaluator, returning a p-value for each feature. Within the feature filter, these scores are combined \cite{benjamini2001control} and the resulting metric is used to directly prune the worst performing features. In this case, we are re-using the evaluator scores within the feature filter, thus there is no additional model training required. ELATE 4o+FRESH outperformed all other non-ELATE baselines in terms of both MAE and RMSE in 4/7 domains and took the least time on 5/7 domains (not including Zero-Shot or T$-$1).

\begin{figure*}[htbp]
  \centering
  \resizebox{0.8\textwidth}{!}{%
    \begin{minipage}{\textwidth}
      \centering
      \lstset{
        style=pythonstyle,
        linewidth=\textwidth
      }
\begin{lstlisting}
# Description
# Store-Specific Weekly Sales Momentum with Temperature Adjustment
# This feature calculates the momentum of weekly sales over the past 4 weeks for each store, adjusted by the average temperature during the same period. By combining sales momentum and temperature, it captures both recent sales trends and environmental factors that might affect consumer behavior, potentially leading to more accurate sales predictions.
df['Store_Specific_Weekly_Sales_Momentum_Temperature_Adjusted'] = df.groupby('Store').apply(lambda x: (x['Weekly_Sales_Tminus1'].rolling(window=4, min_periods=1).mean() - x['Weekly_Sales_Tminus1'].rolling(window=4, min_periods=1).min()) * (1 + 0.01 * x['Temperature'].rolling(window=4, min_periods=1).mean())).reset_index(level=0, drop=True).fillna(0)
feature = df['Store_Specific_Weekly_Sales_Momentum_Temperature_Adjusted']
\end{lstlisting}
      
      \vspace{1em} 
      
\begin{lstlisting}
# Description
# Feature: Accumulation/Distribution Oscillator
# This feature calculates the difference between the Accumulation/Distribution Line (ADL) and its 3-day moving average. The ADL reflects the cumulative flow of money in and out of a security, and the oscillator helps in identifying potential turning points in the price trend by measuring the momentum of money flow. A higher value indicates strong buying or selling pressure.
df['clv'] = ((df['Close'] - df['Low']) - (df['High'] - df['Close'])) / (df['High'] - df['Low'])
df['money_flow_volume'] = df['clv'] * df['Volume']
df['adl'] = df.groupby('Symbol')['money_flow_volume'].cumsum()
df['adl_ma_3'] = df.groupby('Symbol')['adl'].transform(lambda x: x.rolling(window=3).mean())
df['adl_oscillator'] = df['adl'] - df['adl_ma_3']
feature = df['adl_oscillator']
\end{lstlisting}
    \end{minipage}
  }
  \caption{Example feature for the Store (top) and Trading (bottom) datasets. Note: when the code is compiled and executed, only the final \texttt{feature} is returned as a Pandas \cite{mckinney2011pandas} series.}
  \label{fig:example_features_combined}
\end{figure*}

\paragraph{Domain Adaptability}
To highlight the adaptability of ELATE with respect to different tasks, example features are presented for the store and trading datasets in Figure \ref{fig:example_features_combined}.

In the store example, ELATE effectively highlights the influence of both current sales trends and environmental factors on consumer demand. The generated feature combines these elements by calculating a weighted sum of two four-week rolling averages: sales momentum and temperature. The momentum component is derived by taking the difference between the average and minimum weekly sales over the past four weeks. This calculation provides a clear indication of whether sales are on an upward or downward trajectory. An upward trend suggests that sales in the upcoming week are likely to be higher, making it a valuable predictor for future sales performance. The temperature component adjusts the momentum by considering the average temperature over the same period. This adjustment accounts for the likelihood that favorable weather conditions can increase store visits and consumer spending.

In the trading example, ELATE identifies the Accumulation/Distribution Line (ADL) as a key trading indicator. The Close Location Value (CLV) ratio determines the closing price's position between the day's low and high, with a CLV of $-1$ indicating a close at the low and a CLV of $1$ indicating a close at the high. This ratio, multiplied by trading volume, yields the money flow volume, reflecting the net money movement into or out of the asset. A high volume with a positive CLV suggests strong buying interest, potentially driving a price increase. The ADL, calculated as the cumulative sum of money flow volume, tracks overall market interest. Comparing the ADL to its 3-day moving average helps identify momentum shifts: crossing above suggests potential upward momentum, while crossing below indicates potential downward momentum. This indicator is highly specialized and requires expertise in financial mathematics to develop and interpret effectively.

\paragraph{Feature Complexity}
As mentioned previously, one of the challenges with expand and reduce and search based approaches is that they struggle to find complex features composed of many sequential transformations. In the trading example in Figure \ref{fig:example_features_combined}, applying sequential transformations would necessitate a depth of 10:

\begin{lstlisting}[basicstyle=\ttfamily\footnotesize, numbers=left, numberstyle=\tiny, stepnumber=1, numbersep=5pt]
close_low_diff = close - low
high_close_diff = high - close
high_low_diff = high - low
close_loc = close_low_diff - high_close_diff
clv = close_loc / high_low_diff
money_flow_volume = clv * volume
group = groupby(symbol)
adl = group.cumsum(money_flow_volume)
adl_ma_3  = group.rolling_mean(adl, 3)
adl_oscillator = adl - adl_ma_3
\end{lstlisting}

Assuming that these are our only transformations, we have 357 possible transformations at a depth of 1: we can group by 5 columns, we have 110 permutations of the 11 numeric columns for both the subtract and divide operator, 55 combinations of the 11 numeric columns for both the add and multiply operator, and 11 possible numeric columns for which we can compute the cumulative sum and the 3 week rolling mean. This is assuming that we know that the 3 week rolling mean is going to be useful for our problem. Rolling mean is itself composed of numerous base operators and can be computed across a span of any natural numbers. Note, that ELATE does not need to reason over any of these intermediate states and understands given the context that ADL will be useful in stock return prediction. We believe this is a significantly more powerful approach than reasoning over the base transformations explicitly.

\section{Conclusion}\label{sec:conclusion}

In summary, we introduced ELATE, a novel approach to TS Auto-FE. ELATE leverages an LLM to propose contextually useful feature transformations within an evolutionary framework.

Our experimental results demonstrate that ELATE is able to engineer high quality features, resulting in greater forecast accuracy than state-of-the-art TS Auto-FE methods. ELATE is also able to adapt to new forecasting tasks, generating relevant features for a spectrum of applications. Unlike previous expand and reduce based approaches such as \texttt{tsfresh}, ELATE is both time and memory efficient, capable of engineering features for large scale problems (179,568 rows) in a matter of hours. This is achieved by utilizing the extensive knowledge base of LLMs as a heuristic. Unlike deep neural networks which lack interpretability, ELATE explicitly returns the code to generate the features, as well as a comment on their potential utility. This will be particularly beneficial in high risk applications where interpretability is a key concern, for example in healthcare and finance.

We conclude by addressing some limitations. First, the cost of iteratively querying LLMs may be prohibitive for some applications. This could be mitigated by using open source or cheaper language models. We would like to emphasize, that the cost of paying a data scientist to perform this task is likely to be significantly more. It is important to stress, that ELATE is not intended to replace the data-scientist in this task. There is no guarantee that the generated feature transformations are correct and therefore, we firmly believe that ELATE should be used with a human in-the-loop. Skilled data scientists are still required to understand whether the feature transformations make sense for the given task.

Second, ELATE is not guaranteed to reduce the validation set RMSE each generation. While interventional SHAP faithfully captures the importance of individual features used by a predictive model~\cite{chen2020truemodeltruedata}, 
it is plausible that the top $N$ generated features are not being chosen at each step of the ELATE algorithm, but instead a set with overlapping elements. To improve the optimization capability of ELATE, a more sophisticated feature importance metric could be used within the feature selection filter. Finally, in this paper, the same prompt template was used throughout, however future work could focus on using more advanced prompting strategies to encourage better responses \cite{liu2024evolution}.

\paragraph{Disclaimer}

This paper was prepared for informational purposes by the Artificial Intelligence Research group of JPMorgan Chase \& Co. and its affiliates (``JP Morgan'') and is not a product of the Research Department of JP Morgan. JP Morgan makes no representation and warranty whatsoever and disclaims all liability, for the completeness, accuracy or reliability of the information contained herein. This document is not intended as investment research or investment advice, or a recommendation, offer or solicitation for the purchase or sale of any security, financial instrument, financial product or service, or to be used in any way for evaluating the merits of participating in any transaction, and shall not constitute a solicitation under any jurisdiction or to any person, if such solicitation under such jurisdiction or to such person would be unlawful.

\noindent© 2025 JPMorgan Chase \& Co. All rights reserved.

\bibliography{elate}

\newpage
\appendix
\raggedbottom
\onecolumn

\section{Dataset Descriptions}\label{app:description}
\lstset{
  style=text,
  linewidth=\columnwidth
}
\subsection*{ILI}
\begin{lstlisting}
The following contains CDC data on outpatient hospital admissions as a result of influenza-like illnesses in the United States. We have been tasked with developing a time-series prediction model to forecast the percentage of Influenza-like Illness (ILI) visits, adjusted for population differences, in future weeks. This model will help the organization anticipate trends in ILI activity and implement effective measures to mitigate the impact of influenza on communities.

The input features and target to be predicted are:

# INPUT FEATURES

year (int32): The year of the report, indicating the specific week for which the data is collected.
month (int32): The month of the report.
day (int32): The day of the report.
weighted_ili_Tminus1 (float64): The percentage of patient visits to healthcare providers for influenza-like illness (ILI) from the previous week, adjusted for state population differences. This measure provides a standardized view of ILI activity across regions by accounting for variations in population size.
unweighted_ili_Tminus1 (float64): The percentage of patient visits to healthcare providers for influenza-like illness (ILI) from the previous week, not adjusted for population differences.
age_0_4 (int32): The number of ILI visits for patients aged 0 to 4 years.
age_5_24 (int32): The number of ILI visits for patients aged 5 to 24 years.
ILITOTAL (int32): The total number of ILI visits across all age groups.
num_providers (int32): The number of healthcare providers reporting data for the given week.
total_patients (int32): The total number of patients seen by the reporting providers during the given week.

# TARGET

weighted_ili (float64): The percentage of patient visits to healthcare providers for influenza-like illness (ILI), adjusted for state population differences, to be predicted for the next week.
\end{lstlisting}
\label{fig:ili-dataset-description}

\subsection*{Store}
\begin{lstlisting}
One of the leading retail stores in the US, Walmart, would like to predict their sales accurately. Historical sales data for 45 Walmart stores located in different regions is available. The goal is to use features such as economic conditions including CPI and Unemployment Index to predict sales for the coming week.

There are certain events and holidays which impact sales on each day. Walmart runs several promotional markdown events throughout the year. These markdowns precede prominent holidays, the four largest of which are the Super Bowl, Labour Day, Thanksgiving, and Christmas. Part of the challenge presented is modeling the effects of markdowns on these holiday weeks. 

The input features and target to be predicted are:

# INPUT FEATURES

Store (int64): A unique integer identifier for the store.
Holiday_Flag (int64): Boolean flag for whether the week is a special holiday week: 1 : Holiday week 0 : Non-holiday week. Holiday Events include Super Bowl, Labour Day, Thanksgiving and Christmas.
Temperature (float64): Temperature on the day of sale in degrees Fahrenheit.
Fuel_Price (float64): Cost of fuel in the region.
CPI (float64): Prevailing consumer price index.
Unemployment (float64): Prevailing unemployment rate
Weekly_Sales_Tminus1 (float64): Total sales for the given store on that week.
Week (UInt32): Integer representing the week of the year.
Month (int32): Integer representing the month of the year.
Year (int32): Year as an integer.
time_index (int64): Ordered index for the week. This is 1 for the earliest week in the dataset and increases each week.

# TARGET

Weekly_Sales (float64): Total sales for the given store in the coming week.
\end{lstlisting}
\label{fig:store-dataset-description}

\subsection*{ETTh1}
\begin{lstlisting}
This dataset can be used to predict transformers' oil temperature. This helps manage electric power distribution efficiently and prevent waste and equipment depreciation. The input features include various power load metrics, with the oil temperature as the target for prediction.

The input features and target to be predicted are:

# INPUT FEATURES

HUFL (float64): High UseFul Load, representing the high level of useful power load.
HULL (float64): High UseLess Load, representing the high level of useless power load.
MUFL (float64): Middle UseFul Load, representing the middle level of useful power load.
MULL (float64): Middle UseLess Load, representing the middle level of useless power load.
LUFL (float64): Low UseFul Load, representing the low level of useful power load.
LULL (float64): Low UseLess Load, representing the low level of useless power load.
Year (int32): Year.
Month (int32): Month.
Day (int32): Day.
Hour (int32): Hour.
OT_Tminus1 (float64): Current oil temperature indicating the condition of the electrical transformer.

# TARGET

OT (float64): Oil Temperature one hour in advance.
\end{lstlisting}
\label{fig:ett-dataset-description}

\subsection*{Energy}
\begin{lstlisting}
This dataset contains 4 years of electrical consumption, generation, pricing, and weather data for Spain. Consumption and generation data was retrieved from ENTSOE a public portal for Transmission Service Operator (TSO) data. Settlement prices were obtained from the Spanish TSO Red Electric Espa\~na. Weather data was extracted from Open Weather API. Each row represents a particular hour, while the target to be predicted is the total energy demand 24 hours in the future.

The input features and target to be predicted are:

# INPUT FEATURES

total_generation (float64): Total energy generation that hour in MW.
total_load_forecast (float64): Forecasted electrical energy demand that hour in MW.
total_load_actual (float64): Actual electrical energy demand that hour in MW.
price_day_ahead (float64): Forecasted energy price for that day in EUR/MWh.
price_actual (float64): Actual energy price for that dat in EUR/MWh.
temp (float64): Average temperature in K
temp_min (float64): Minimum temperature in K
temp_max (float64): Maximum temperature in K
pressure (float64): Pressure in hPa
humidity (float64): Humidity in %
wind_speed (float64): Wind speed in m/s
wind_deg (float64): Wind direction in degrees.
rain_1h (float64): Rain in the last hour in mm.
rain_3h (float64): Rain in the last 3 hours in mm.
snow_3h (float64): Snow in the last 3 hours in mm.
clouds_all (float64): Cloud cover in %.
year (int64): Year.
month (int64): Month.
day (int64): Day.
hour (int64): Hour.

# TARGET

target (float64): Total electrical energy demand 24 hours in advance in MW.
\end{lstlisting}
\label{fig:energy-dataset-description}

\subsection*{Pollution}
\begin{lstlisting}
This is a dataset that reports on the weather and the level of pm2.5 pollution each hour for five years at the US embassy in Beijing, China.

The data includes the date and time information, the current pollution concentration, and the weather information including dew point, temperature, pressure, wind direction, wind speed and the cumulative number of hours of snow and rain. The target to be predicted, and the input features are as follows:

# INPUT FEATURES

time (int64): Number of hours since the start of record.
hour (int64): Hour of the day (between 0 and 23)
day (int64): Day of the month (between 1 and 31)
month (int64): Month of the year (between 1 and 12)
year (int64): Year
pollution_Tminus1 (float64): Current pollution concentration.
dew (int64): Dew Point
temp (float64): Temperature
press (float64): Pressure
wnd_dir (category): Combined wind direction
wnd_spd (float): Cumulated wind speed
snow (int64): Cumulated hours of snow
rain (int64): Cumulated hours of rain

# TARGET

pollution (float64): Pollution concentration for the same time the next day.

We can use this data and frame a forecasting problem where, given the weather conditions and current pollution concentration, we forecast the pollution for the same time the next day.
\end{lstlisting}
\label{fig:pollution-dataset-description}

\subsection*{Trading}
\begin{lstlisting}
The Standard and Poor's 500 or S&P 500 is the most famous financial benchmark in the world. This stock market index tracks the performance of 500 large companies listed on stock exchanges in the United States. As of December 31, 2020, more than $5.4 trillion was invested in assets tied to the performance of this index.

You have access to daily price data for each stock in the index, as well as the daily value of the index. Your goal is to predict the excess returns of the stock over the next day in relation to the index.

The input features and target to be predicted are:

# INPUT FEATURES

Symbol (category): String "Ticker". Unique code given to the company listed on the exchange.
Close (float64): Price at market closure that trading day.
High (float64): Maximum price that trading day.
Low (float64): Minimum price that trading day.
Open (float64): Price at market opening that trading day.
Volume (float64): Volume traded that day.
S&P500 (float64): Value of the S&P 500 index on that day.
Year (int32): Year of the trading date.
Month (int32): Month of the trading date. 
Day (int32): Day of the month trading date.
Target_Tminus1 (float64): The excess returns of the stock over the previous day in relation to the index. Lagged value of the target defined below.
Time_Index (int32): Ordered index for the date. This is 1 for the earliest date in the dataset and increases each day.

# TARGET

Target (float64): The excess returns of the stock over the next day in relation to the index. Return for the stock is computed using the close price. The unit of the target is basis points, which is a common unit of measurement in financial markets. A 1 basis point price move is equivalent to a 0.01% price move.
Where t is the time at the current price and t+1 is the next days price, we can define the target:
$$
target = (\frac{close_{t+1} - close_{t}}{close_{t}} - \frac{S&P500_{t+1} - S&P500_{t}}{S&P500_{t}}) * 10000
$$

All prices are in USD.
\end{lstlisting}
\label{fig:trading-dataset-description}

\subsection*{Food}
\begin{lstlisting}
There is a meal delivery company which operates in multiple cities. They have various fulfilment centers in these cities for dispatching meal orders to their customers. The client wants you to help these centers with demand forecasting so that these centers can plan the stock of raw materials accordingly. The replenishment of raw materials is done on a weekly basis and since the raw material is perishable, the procurement planning is of utmost importance.

The data includes the date information, current order numbers, information related to the meal type and fullfilment center as well as whether or not any promotion was running. Given this, we must forecast the number of orders required for the next week. The target to be predicted, and the input features are as follows:

# INPUT FEATURES

week (int64): Integer encoding of the week. Note that this is just a sequential encoding and does not represent the actual week of the year. This column can have more than 52 values.
center_id (int64): Integer identifier of the fulfilment center.
meal_id (int64): Integer identifier of the meal.
checkout_price (float64): Checkout price of the meal. What is paid after promotions have been applied.
base_price (float64): Base price of the meal. Price before any promotion has been applied.
emailer_for_promotion (int64): Boolean representing whether an email has been sent about a promotion for the meal. 1 is True, 0 is False.
homepage_featured (int64): Boolean representing whether the meal is featured on the companies home page. 1 is True, 0 is False.
center_type (category): Categorical representing the type of fulfilment center. Can be: TYPE_B, TYPE_A, TYPE_C.
op_area (float64): Float describing the operational area that the fulfilment center serves.
num_orders_Tminus1 (float64): Total number of orders for the meal and fulfilment center for the current week.

# TARGET

num_orders (int64): The total number of orders for the meal and fulfilment center for the next week.
\end{lstlisting}
\label{fig:food-dataset-description}

\newpage
\section{Example Features}\label{app:examples}
Here we display the top 10 generated features for each domain.
\subsection*{ILI}
\lstset{
  style=pythonstyle,
  linewidth=\columnwidth
}
\begin{lstlisting}
# Description
# The exponential weighted moving average of the 'weighted_ili_Tminus1' over a 1-week window.
# This feature gives more importance to the most recent week's data while discounting older weeks, which might help in capturing the immediate past trends in ILI activity and improve prediction accuracy.
df['weighted_ili_Tminus1_ewm_1w'] = df['weighted_ili_Tminus1'].ewm(span=1, adjust=False).mean().fillna(0)
feature = df['weighted_ili_Tminus1_ewm_1w']


# Description
# This feature calculates the ratio of the exponentially weighted moving average (EWMA) of 'weighted_ili_Tminus1' over a 1-week window to the EWMA of 'unweighted_ili_Tminus1' over a 2-week window. By capturing the relationship between the weighted and unweighted ILI percentages with different smoothing spans, this feature can help identify discrepancies and trends in ILI reporting across different states, potentially indicating areas with underreporting or overreporting, which can be valuable for predicting shifts in weighted ILI percentages.
df['weighted_to_unweighted_ili_Tminus1_ewm_ratio_1w_2w'] = (df['weighted_ili_Tminus1'].ewm(span=1, adjust=False).mean() / df['unweighted_ili_Tminus1'].ewm(span=2, adjust=False).mean()).fillna(0)
feature = df['weighted_to_unweighted_ili_Tminus1_ewm_ratio_1w_2w']


# Description
# Month of the year as a cyclical feature
# This helps the model understand the cyclical nature of months in a year.
df['month_sin'] = np.sin(2 * np.pi * df['month'] / 12)
feature = df['month_sin']


# Description
# This feature computes the ratio of the exponential weighted moving average (EWMA) of 'weighted_ili_Tminus1' over a 4-week window to the EWMA of 'ILITOTAL' over a 2-week window. This captures a more stable trend in weighted ILI visits in relation to the recent overall ILI activity, potentially improving predictions by considering both mid-term specific and recent general ILI patterns.
df['weighted_ili_Tminus1_ewm_4w_to_ILITOTAL_ewm_2w_ratio'] = df['weighted_ili_Tminus1'].ewm(span=4, adjust=False).mean() / df['ILITOTAL'].ewm(span=2, adjust=False).mean().replace(0, np.nan)
feature = df['weighted_ili_Tminus1_ewm_4w_to_ILITOTAL_ewm_2w_ratio']


# Description
# Creating a feature that captures the ratio of ILI visits for patients aged 5 to 24 years to the total number of ILI visits. This feature emphasizes the contribution of the school-age and young adult population to the overall ILI activity, which can be significant due to their social interactions and mobility patterns. Understanding this demographic's impact on ILI rates can aid in predicting future trends.
df['age_5_24_to_total_ratio'] = df['age_5_24'] / df['ILITOTAL']
df['age_5_24_to_total_ratio'] = df['age_5_24_to_total_ratio'].fillna(0)
feature = df['age_5_24_to_total_ratio']


# Description
# This feature computes the ratio of the exponential weighted moving average (EWMA) of 'weighted_ili_Tminus1' over a 1-week window to the EWMA of 'ILITOTAL' over a 3-week window. This captures the recent trend in weighted ILI visits in relation to the overall ILI activity, potentially improving predictions by considering both recent specific and general ILI patterns.
df['weighted_ili_Tminus1_ewm_1w_to_ILITOTAL_ewm_3w_ratio'] = df['weighted_ili_Tminus1'].ewm(span=1, adjust=False).mean() / df['ILITOTAL'].ewm(span=3, adjust=False).mean().replace(0, np.nan)
feature = df['weighted_ili_Tminus1_ewm_1w_to_ILITOTAL_ewm_3w_ratio']


# Description
# Creating a feature that calculates the rolling standard deviation of 'weighted_ili_Tminus1' over a 2-week window. This feature captures the variability in ILI activity over recent weeks, helping to identify periods of stability or volatility in flu trends. Such information can be crucial for predicting abrupt changes in ILI rates.
df['weighted_ili_Tminus1_rolling_std_2w'] = df['weighted_ili_Tminus1'].rolling(window=2).std().fillna(0)
feature = df['weighted_ili_Tminus1_rolling_std_2w']


# Description
# This feature computes the exponentially weighted moving average (EWMA) of the 'unweighted_ili_Tminus1' over a 1-week span and divides it by the 'ILITOTAL' over the same period. This captures the relationship between the unweighted ILI percentage from the previous week and the total ILI visits, providing insight into how past ILI percentages relate to the overall burden of ILI visits, which can be valuable for predicting future ILI activity.
df['unweighted_ili_Tminus1_ewm_1w_to_ILITOTAL_ratio'] = (df['unweighted_ili_Tminus1'].ewm(span=1, adjust=False).mean() / df['ILITOTAL']).fillna(0)
feature = df['unweighted_ili_Tminus1_ewm_1w_to_ILITOTAL_ratio']


# Description
# This feature calculates the ratio of the 'unweighted_ili_Tminus1' to the square root of the sum of 'age_0_4' and 'age_5_24' over a 3-week rolling window. By examining the relationship between past ILI percentages and the sum of young age ILI visits, this feature captures the influence of youth ILI activity on overall ILI trends, which can be critical in anticipating future ILI visits since younger populations often contribute significantly to ILI spread.
df['unweighted_ili_Tminus1_rolling_3w_per_sqrt_young_age_sum'] = (df['unweighted_ili_Tminus1'] / np.sqrt((df['age_0_4'] + df['age_5_24']).rolling(window=3).sum())).fillna(0)
feature = df['unweighted_ili_Tminus1_rolling_3w_per_sqrt_young_age_sum']


# Description
# Calculating the rolling standard deviation of the 'ILITOTAL' over a 2-week window. This feature captures the volatility in the total number of ILI visits over the past two weeks, which might provide insights into sudden changes or spikes in ILI activity, aiding in predicting future ILI percentages by identifying patterns of increased variability.
df['ILITOTAL_rolling_std_2w'] = df['ILITOTAL'].rolling(window=2).std().fillna(0)
feature = df['ILITOTAL_rolling_std_2w']
\end{lstlisting}
\label{fig:ili-best-features}

\subsection*{Store}
\lstset{
  style=pythonstyle,
  linewidth=\columnwidth
}
\begin{lstlisting}
# Description:
# Rolling Sales Mean
# The average weekly sales over the past 4 weeks for each store. This feature helps to smooth out short-term fluctuations and highlight longer-term trends or cycles in sales, which can be useful in predicting future sales.
df['Rolling_Sales_Mean'] = df.groupby('Store')['Weekly_Sales_Tminus1'].transform(lambda x: x.shift().rolling(window=4, min_periods=1).mean())
feature = df['Rolling_Sales_Mean']


# Description:
# Rolling Sales Variance
# The variance of weekly sales over the past 4 weeks for each store. This feature captures the volatility in sales, which can provide insight into fluctuating sales patterns and help improve prediction accuracy.
df['Rolling_Sales_Variance'] = df.groupby('Store')['Weekly_Sales_Tminus1'].transform(lambda x: x.rolling(window=4, min_periods=1).var())
feature = df['Rolling_Sales_Variance']


# Description
# Temperature Influence on Sales Variability
# This feature captures the variability of sales in relation to temperature changes over the past four weeks. By calculating the rolling standard deviation of sales influenced by temperature, it helps to identify periods during which temperature fluctuations might have caused significant variations in sales, thereby providing insights into consumer behavior patterns related to weather conditions.
df['Sales_Temperature_Variability'] = df.groupby('Store')['Weekly_Sales_Tminus1'].transform(lambda x: x.rolling(window=4, min_periods=1).std() * df['Temperature'])
feature = df['Sales_Temperature_Variability']


# Description
# Store-Specific Rolling Sales Variance
# This feature calculates the rolling variance of weekly sales over the past 8 weeks for each store. It captures the variability in sales, which can indicate the stability or volatility of sales patterns for a store. Understanding sales variance can help in predicting future sales trends and adjusting strategies accordingly.
df['Store_Specific_Rolling_Sales_Variance'] = df.groupby('Store')['Weekly_Sales_Tminus1'].transform(lambda x: x.shift().rolling(window=8, min_periods=1).var().fillna(0))
feature = df['Store_Specific_Rolling_Sales_Variance']


# Description
# Store-Specific Rolling Holiday Adjusted Sales Variance
# This feature calculates the rolling variance of weekly sales for each store over the past 8 weeks, excluding holiday weeks. By focusing on non-holiday periods, it captures the typical sales volatility, providing insights into the store's regular sales fluctuations and helping to distinguish the impact of promotional events from normal sales patterns.
df['Store_Specific_Rolling_Holiday_Adjusted_Sales_Variance'] = df.groupby('Store').apply(lambda x: x.loc[x['Holiday_Flag'] == 0, 'Weekly_Sales_Tminus1'].shift().rolling(window=8, min_periods=1).var()).reset_index(level=0, drop=True).fillna(0)
feature = df['Store_Specific_Rolling_Holiday_Adjusted_Sales_Variance']


# Description
# Store-Specific Monthly Sales Seasonality
# This feature calculates the average weekly sales for each store in the same month over the past years. By capturing the store-specific sales pattern for each month, the model can leverage historical seasonality trends to improve predictions.
df['Store_Specific_Monthly_Sales_Seasonality'] = df.groupby(['Store', 'Month'])['Weekly_Sales_Tminus1'].transform(lambda x: x.expanding().mean().shift().fillna(0))
feature = df['Store_Specific_Monthly_Sales_Seasonality']


# Description
# Store-Specific Rolling Weekly Sales Standard Deviation
# This feature calculates the rolling standard deviation of weekly sales for each store over the past 6 weeks. It provides insight into the variability or volatility of sales at the store level, which can be indicative of factors such as market dynamics, promotional effects, or seasonal variations. Understanding sales volatility allows the model to adjust its predictions to account for periods of instability or consistent sales, potentially improving forecast accuracy.
df['Store_Specific_Rolling_Weekly_Sales_StdDev'] = df.groupby('Store')['Weekly_Sales_Tminus1'].transform(lambda x: x.shift().rolling(window=6, min_periods=1).std().fillna(0))
feature = df['Store_Specific_Rolling_Weekly_Sales_StdDev']


# Description
# Store-Specific Weekly Sales Momentum with Temperature Adjustment
# This feature calculates the momentum of weekly sales over the past 4 weeks for each store, adjusted by the average temperature during the same period. By combining sales momentum and temperature, it captures both recent sales trends and environmental factors that might affect consumer behavior, potentially leading to more accurate sales predictions.
df['Store_Specific_Weekly_Sales_Momentum_Temperature_Adjusted'] = df.groupby('Store').apply(lambda x: (x['Weekly_Sales_Tminus1'].rolling(window=4, min_periods=1).mean() - x['Weekly_Sales_Tminus1'].rolling(window=4, min_periods=1).min()) * (1 + 0.01 * x['Temperature'].rolling(window=4, min_periods=1).mean())).reset_index(level=0, drop=True).fillna(0)
feature = df['Store_Specific_Weekly_Sales_Momentum_Temperature_Adjusted']


# Description
# Store-Specific Rolling Sales Temperature Interaction Mean
# This feature calculates the rolling mean of the interaction between weekly sales and temperature for each store over a 3-week window. By capturing the average influence of temperature on sales over time, it helps to identify patterns where temperature changes might have a consistent impact on sales, which can be useful for forecasting future sales.
df['Store_Specific_Rolling_Sales_Temperature_Interaction_Mean'] = df.groupby('Store').apply(lambda x: (x['Weekly_Sales_Tminus1'] * x['Temperature']).rolling(window=3, min_periods=1).mean()).reset_index(level=0, drop=True)
feature = df['Store_Specific_Rolling_Sales_Temperature_Interaction_Mean']


# Description
# Store-Specific Rolling Holiday Sales Mean
# This feature calculates the rolling mean of sales during holiday weeks for each store over the past 4 holiday weeks. It helps to understand the average sales performance of a store during holiday periods, providing insights into the store's sales trend during these high-impact weeks.
df['Store_Specific_Rolling_Holiday_Sales_Mean'] = df.groupby('Store').apply(lambda x: x.loc[x['Holiday_Flag'] == 1, 'Weekly_Sales_Tminus1'].rolling(window=4, min_periods=1).mean()).reset_index(level=0, drop=True).fillna(0)
feature = df['Store_Specific_Rolling_Holiday_Sales_Mean']
\end{lstlisting}
\label{fig:store-best-features}

\subsection*{ETTh1}
\lstset{
  style=pythonstyle,
  linewidth=\columnwidth
}
\begin{lstlisting}
# Description
# This feature calculates the exponential weighted moving average (EWMA) of the transformer oil temperature from the previous hour 
# (OT_Tminus1) over a 9-hour window. It helps capture the recent trends and smoothens the fluctuations in oil temperature, providing 
# a more stable and predictive feature for modeling the future temperature, allowing the model to learn from recent patterns efficiently.
df['OT_Tminus1_ewma_9h'] = df['OT_Tminus1'].ewm(span=9, adjust=False).mean().fillna(method='ffill')
feature = df['OT_Tminus1_ewma_9h']


# Description
# This feature calculates the exponential weighted moving average (EWMA) of the interaction between the previous hour's oil temperature (OT_Tminus1) and the low useful load (LUFL) over the past 18 hours. The EWMA provides a smoothed time series that emphasizes recent interactions between OT_Tminus1 and LUFL, capturing potential relationships between temperature changes and low useful load levels, which may help in predicting future oil temperature trends.
df['OT_Tminus1_LUFL_interaction_ewma_18h'] = ((df['OT_Tminus1'] * df['LUFL']).ewm(span=18, adjust=False).mean()).fillna(0)
feature = df['OT_Tminus1_LUFL_interaction_ewma_18h']


# Description
# This feature calculates the exponential weighted moving average (EWMA) of the ratio of high useful load (HUFL) to the previous oil temperature (OT_Tminus1) over a 12-hour window.
# This feature aims to capture the impact of high useful load on the transformer oil temperature, smoothing out short-term fluctuations while emphasizing recent changes. 
# It's useful for understanding how variations in useful load might influence temperature trends and thereby inform predictions.
df['HUFL_OT_Tminus1_ratio_ewma_12h'] = (df['HUFL'] / df['OT_Tminus1'].replace(0, np.nan)).ewm(span=12, adjust=False).mean().fillna(method='ffill')
feature = df['HUFL_OT_Tminus1_ratio_ewma_12h']


# Description
# This feature calculates the exponentially weighted moving average (EWMA) of the interaction between the current oil temperature (OT_Tminus1) and the low level of useful load (LUFL) over a 720-hour window. The interaction captures the influence of the low useful load on the oil temperature, highlighting periods where low useful load is significant. This might help in identifying patterns that affect temperature fluctuations, especially if low useful load has a prolonged impact on the transformer oil temperature.
df['OT_Tminus1_LUFL_interaction_ewma_720h'] = (df['OT_Tminus1'] * df['LUFL']).ewm(span=720, adjust=False).mean().fillna(0)
feature = df['OT_Tminus1_LUFL_interaction_ewma_720h']


# Description
# This feature calculates the standardized deviation of the oil temperature from the previous hour (OT_Tminus1) over a 12-hour window.
# It provides a measure of how much the oil temperature deviates from its mean relative to its standard deviation, capturing fluctuations in temperature.
# This can help in identifying periods of instability or sudden changes in the transformer condition, which may be crucial for predicting future temperatures.
df['OT_Tminus1_standardized_deviation_12h'] = (df['OT_Tminus1'] - df['OT_Tminus1'].rolling(window=12, min_periods=1).mean()) / df['OT_Tminus1'].rolling(window=12, min_periods=1).std().replace(0, np.nan).fillna(0)
feature = df['OT_Tminus1_standardized_deviation_12h']


# Description
# This feature calculates the exponentially weighted moving average (EWMA) of the interaction between the current oil temperature (OT_Tminus1) and the low level of useless load (LULL) over an 840-hour window. This interaction helps capture the prolonged influence of low useless load on the transformer oil temperature, revealing patterns where this type of load may have significant effects on temperature fluctuations. It offers potential insights into the impact of low load inefficiencies on temperature predictions.
df['OT_Tminus1_LULL_interaction_ewma_840h'] = (df['OT_Tminus1'] * df['LULL']).ewm(span=840, adjust=False).mean().fillna(0)
feature = df['OT_Tminus1_LULL_interaction_ewma_840h']


# Description
# This feature calculates the exponential weighted moving average (EWMA) of the OT_Tminus1 over a 12-hour window, adjusted by the ratio of High UseFul Load (HUFL) to the combined High UseFul and High UseLess Load. This adjustment helps capture the impact of high useful load on temperature trends, reflecting how efficiently the load is being utilized and its potential influence on temperature fluctuation.
df['OT_Tminus1_HUFL_adjusted_ewma_12h'] = df['OT_Tminus1'].ewm(span=12, adjust=False).mean() * (df['HUFL'] / (df['HUFL'] + df['HULL']))
feature = df['OT_Tminus1_HUFL_adjusted_ewma_12h']


# Description
# This feature calculates the standardized deviation of the current oil temperature (OT_Tminus1) over a 24-hour window.
# Standardizing the deviation helps in understanding how much the oil temperature deviates from its mean in the past 24 hours, relative to its standard deviation.
# This feature captures both the volatility and significance of temperature changes, which can be critical for predicting the future temperature.
mean_24h = df['OT_Tminus1'].rolling(window=24, min_periods=1).mean()
std_24h = df['OT_Tminus1'].rolling(window=24, min_periods=1).std().fillna(0)
df['OT_Tminus1_standardized_deviation_24h'] = ((df['OT_Tminus1'] - mean_24h) / std_24h).fillna(0)
feature = df['OT_Tminus1_standardized_deviation_24h']


# Description
# This feature calculates the exponential weighted moving average (EWMA) of the interaction between the current oil temperature (OT_Tminus1) and the low useless load (LULL) over a 15000-hour window. This captures long-term trends and variations in the interaction between oil temperature and low useless load, providing insights into how sustained low useless load levels impact the future oil temperature.
df['OT_Tminus1_LULL_interaction_ewma_15000h'] = (df['OT_Tminus1'] * df['LULL']).ewm(span=15000, adjust=False).mean()
feature = df['OT_Tminus1_LULL_interaction_ewma_15000h']


# Description
# This feature calculates the exponentially weighted moving average (EWMA) of the interaction between the oil temperature at the previous hour (OT_Tminus1) and the middle useful load (MUFL) over the past 6 hours.
# This interaction captures the influence of the middle useful load on oil temperature changes, emphasizing periods with higher middle useful load, which may affect future temperature patterns.
df['OT_Tminus1_MUFL_interaction_ewma_6h'] = (df['OT_Tminus1'] * df['MUFL']).ewm(span=6, adjust=False).mean().fillna(0)
feature = df['OT_Tminus1_MUFL_interaction_ewma_6h']
\end{lstlisting}
\label{fig:ett-best-features}

\subsection*{Energy}
\lstset{
  style=pythonstyle,
  linewidth=\columnwidth
}
\begin{lstlisting}
# Description:
# Temperature and Load Forecast Error Interaction with Day of Week
# This feature captures the interaction between temperature and the error in load forecasting, modulated by the day of the week.
# This might be useful in understanding how temperature impacts demand forecasting errors, considering the influence of weekdays and weekends on energy consumption patterns.
df['day_of_week'] = pd.to_datetime(df[['year', 'month', 'day']]).dt.dayofweek
df['temp_load_forecast_error'] = df['total_load_forecast'] - df['total_load_actual']
df['day_of_week_sin'] = np.sin(2 * np.pi * df['day_of_week'] / 7)
df['temp_load_forecast_error_day_interaction'] = df['temp'] * df['temp_load_forecast_error'] * df['day_of_week_sin']
feature = df['temp_load_forecast_error_day_interaction']


# Description
# Temperature and Humidity Load Interaction: This feature captures the interaction between temperature, humidity, and actual electrical energy demand.
# It is designed to reflect how combined weather conditions influence energy consumption, which can be important for accurately predicting future demand.
df['temp_humidity_load_interaction'] = (df['temp'] * df['humidity']) / df['total_load_actual'].replace(0, np.nan).fillna(method='ffill')
feature = df['temp_humidity_load_interaction']


# Description
# Generation and Load Forecast Error Interaction: This feature captures the interaction between the ratio of total generation to actual load over the past 24 hours and the load forecast error (difference between forecasted and actual load). This interaction can reveal how discrepancies between forecasted and actual energy demand relate to generation dynamics, potentially highlighting inefficiencies or adaptive responses in generation capacity.
df['generation_to_load_ratio'] = df['total_generation'] / df['total_load_actual'].replace(0, np.nan).fillna(method='ffill')
df['load_forecast_error'] = df['total_load_forecast'] - df['total_load_actual']
df['generation_load_forecast_error_interaction'] = df['generation_to_load_ratio'].rolling(window=24, min_periods=1).mean().fillna(method='ffill') * df['load_forecast_error']
feature = df['generation_load_forecast_error_interaction']


# Description
# Temperature and Load Interaction
# This feature calculates the interaction between the actual electrical energy demand and the average temperature. Understanding how temperature influences energy demand could help in predicting future demand, as temperature variations are known to impact energy consumption patterns due to heating and cooling needs.
df['temp_load_interaction'] = df['total_load_actual'] * df['temp']
feature = df['temp_load_interaction']


# Description:
# Hourly Sinusoidal Transformation
# A transformation of the hour to capture cyclical daily patterns.
df["hour_sin"] = np.sin(2 * np.pi * df["hour"] / 24)
feature = df['hour_sin']


# Description
# Hourly Diurnal Variation Interaction with Humidity
# This feature captures the interaction between the sinusoidal representation of the hour (to model daily cyclical patterns) and humidity levels. It can be useful in identifying how daily humidity variations interact with cyclical hourly energy demand changes, potentially impacting forecast accuracy.
df['hour_cos'] = np.cos(2 * np.pi * df['hour'] / 24)
df['hour_humidity_interaction'] = df['hour_cos'] * df['humidity']
feature = df['hour_humidity_interaction']


# Description:
# Hourly Temperature Anomaly
# This feature calculates the deviation of the average temperature from the mean temperature of the same hour across the dataset. It might be useful for capturing unusual temperature patterns that could influence energy demand, as deviations from typical temperatures can affect heating or cooling needs.
hourly_mean_temp = df.groupby('hour')['temp'].transform('mean')
df['temp_anomaly'] = df['temp'] - hourly_mean_temp
feature = df['temp_anomaly']


# Description
# Pressure and Temperature Interaction Rolling Mean: This feature captures the interaction between atmospheric pressure and temperature over the past 48 hours.
# This interaction might reflect the combined effect of pressure and temperature on energy demand, as both factors can influence heating and cooling requirements.
df['pressure_temp_interaction_rolling_mean'] = (df['pressure'] * df['temp']).rolling(window=48, min_periods=1).mean().fillna(method='ffill')
feature = df['pressure_temp_interaction_rolling_mean']


# Description
# Generation Efficiency Weighted by Wind Speed and Temperature: This feature calculates the efficiency of energy generation by considering both wind speed and temperature over the past 24 hours. It aims to capture the influence of weather conditions on how effectively energy is generated. Higher wind speeds can enhance wind energy generation, while temperature variations can affect solar and thermal generation efficiency. By weighting generation efficiency with these factors, this feature provides insights into how weather conditions impact energy generation capabilities, which can help improve predictions of future energy demand.
df['generation_efficiency'] = df['total_generation'] / df['total_load_actual']
df['weighted_generation_efficiency'] = df['generation_efficiency'] * (df['wind_speed'].rolling(window=24, min_periods=1).mean().fillna(method='ffill') + df['temp'].rolling(window=24, min_periods=1).mean().fillna(method='ffill'))
feature = df['weighted_generation_efficiency']


# Description:
# Rolling Mean of Load Generation Ratio
# This feature calculates the rolling mean of the ratio between total energy generation and actual load over the past 24 hours,
# which could be useful for understanding the balance between supply and demand and identifying trends in energy production versus consumption.
df['load_generation_ratio'] = df['total_generation'] / (df['total_load_actual'] + 1e-9)  # Adding a small constant to avoid division by zero
df['load_generation_ratio_rolling_mean'] = df['load_generation_ratio'].rolling(window=24, min_periods=1).mean().fillna(method='ffill')
feature = df['load_generation_ratio_rolling_mean']
\end{lstlisting}
\label{fig:energy-best-features}

\subsection*{Pollution}
\lstset{
  style=pythonstyle,
  linewidth=\columnwidth
}
\begin{lstlisting}
# Description
# Rolling standard deviation of temperature over the last 12 hours
# This feature can help the model capture variability in temperature, which might be a factor influencing pollution levels.
df['temp_rolling_std_12h'] = df['temp'].rolling(window=12).std().fillna(method='ffill')
feature = df['temp_rolling_std_12h']


# Description
# This feature computes the exponential weighted moving average (EWMA) of the pressure over the past 72 hours. Pressure influences weather patterns and can affect air quality by influencing pollutant dispersion. This feature helps capture trends and smooth fluctuations in pressure which might contribute to pollution levels.
df['press_ewma_72h'] = df['press'].ewm(span=72, adjust=False).mean().fillna(method='ffill')
feature = df['press_ewma_72h']


# Description
# The 72-hour rolling average of the temperature to pollution ratio. This feature captures the mean relationship between temperature and pollution levels over the past three days, potentially indicating how temperature variations impact pollution dispersion or accumulation, which can be useful for predicting future pollution levels.
df['temp_pollution_ratio_rolling_mean_72h'] = (df['temp'] / df['pollution_Tminus1']).rolling(window=72).mean().fillna(method='ffill')
feature = df['temp_pollution_ratio_rolling_mean_72h']


# Description
# Exponential moving average of pollution concentration over a 3-hour window
# This feature captures recent trends in pollution levels, providing insight into short-term fluctuations that may impact the next day's pollution prediction.
df['pollution_ema_3h'] = df['pollution_Tminus1'].ewm(span=3, adjust=False).mean()
feature = df['pollution_ema_3h']


# Description
# The 72-hour exponentially weighted moving average (EWMA) of the pollution to temperature ratio. This feature captures the recent trend in the relationship between pollution and temperature, which can be indicative of atmospheric conditions affecting pollution dispersal and concentration.
df['pollution_temp_ratio_ewma_72h'] = (df['pollution_Tminus1'] / df['temp']).ewm(span=72, adjust=False).mean().fillna(method='ffill')
feature = df['pollution_temp_ratio_ewma_72h']


# Description
# The 48-hour rolling maximum of the dew point to temperature ratio provides insights into the highest relative humidity levels observed over a two-day period. This feature captures extreme conditions of humidity which might affect pollution levels by influencing atmospheric composition and dispersion patterns.
df['dew_temp_ratio_rolling_max_48h'] = (df['dew'] / df['temp']).rolling(window=48).max().fillna(method='ffill')
feature = df['dew_temp_ratio_rolling_max_48h']


# Description
# Cosine transformation of the hour of the day to capture daily cyclical patterns in pollution levels. This feature can help in identifying time-of-day effects on pollution, which are often cyclical due to human activities and atmospheric conditions.
df['hour_cos'] = np.cos(2 * np.pi * df['hour'] / 24)
feature = df['hour_cos']


# Description
# The 72-hour rolling sum of the cumulated wind speed. This feature captures the total wind activity over the past three days, which can be useful for predicting pollution dispersal patterns as wind speed influences the movement and dilution of pollutants in the atmosphere.
df['wnd_spd_rolling_sum_72h'] = df['wnd_spd'].rolling(window=72).sum().fillna(method='ffill')
feature = df['wnd_spd_rolling_sum_72h']


# Description
# The ratio of dew point to temperature averaged over the past 24 hours. This feature captures the relative humidity trend, which can influence pollution levels by affecting the dispersion and chemical transformation of pollutants in the air.
df['dew_temp_ratio_24h'] = (df['dew'].rolling(window=24, min_periods=1).mean() / df['temp'].rolling(window=24, min_periods=1).mean()).fillna(method='ffill')
feature = df['dew_temp_ratio_24h']


# Description
# The 72-hour rolling maximum of the dew to pollution ratio. This feature captures the peak relationship between dew point and pollution levels over the past three days, which can indicate periods when the dew point is high relative to pollution levels, potentially influencing future pollution predictions due to atmospheric conditions that affect pollutant dispersion or accumulation.
df['dew_pollution_ratio_rolling_max_72h'] = (df['dew'] / df['pollution_Tminus1']).rolling(window=72).max().fillna(method='ffill')
feature = df['dew_pollution_ratio_rolling_max_72h']
\end{lstlisting}
\label{fig:pollution-best-features}

\subsection*{Trading}
\lstset{
  style=pythonstyle,
  linewidth=\columnwidth
}
\lstinputlisting{trading_best_features.txt}
\label{fig:trading-best-features}

\subsection*{Food}
\lstset{
  style=pythonstyle,
  linewidth=\columnwidth
}
\lstinputlisting{food_best_features.txt}
\label{fig:food-best-features}
\newpage

\newpage

\section{Implementation Details}

\subsection*{Feature}
We use a custom Feature object to store each generated feature. To enable re-computation with other datasets, we store the code to generate the feature as opposed to the feature vector itself. Algorithm \ref{alg:add_scores} stores the scores for each evaluator to the Feature object. Algorithm \ref{alg:get_feature} first checks whether the code is allowable using AST parsing and if so, it compiles and executes the code, resulting in the feature vector $X_j$.

\begin{algorithm}
\caption{Constructor Method of Feature Class}
\label{alg:feat_constructor}
\begin{algorithmic}
\STATE \textbf{Input:}
\STATE \hspace{1em} $self$: Feature object
\STATE \hspace{1em} $code$: String code for generating feature
\STATE \textbf{Output:}
\STATE \hspace{1em} Initialized Feature object
\STATE $self.code \leftarrow code$
\STATE $self.name \leftarrow \text{None}$
\STATE $self.score \leftarrow \text{None}$
\STATE $self.scores \leftarrow \text{None}$
\end{algorithmic}
\end{algorithm}

\begin{algorithm}
\caption{add\_score Method of Feature Class}
\label{alg:add_scores}
\begin{algorithmic}
\STATE \textbf{Input:}
\STATE \hspace{1em} $self$: Feature object
\STATE \hspace{1em} $scores$: Dictionary of scores from evaluators
\STATE \textbf{Output:}
\STATE \hspace{1em} Updated Feature object with scores
\STATE $self.scores \leftarrow scores$
\STATE $self.score \leftarrow \text{mean}(scores)$
\end{algorithmic}
\end{algorithm}

\begin{algorithm}
\caption{get\_feature Method of Feature Class}
\label{alg:get_feature}
\begin{algorithmic}
\STATE \textbf{Input:}
\STATE \hspace{1em} $self$: Feature object
\STATE \hspace{1em} $X$: Feature matrix dataframe
\STATE \textbf{Output:}
\STATE \hspace{1em} Feature series
\STATE $parsed\_code \leftarrow \text{ast\_parse}(self.code)$
\IF{$\text{is\_valid}(parsed\_code)$}
    \STATE $compiled\_code \leftarrow \text{compile}(parsed\_code)$
    \STATE $vars \leftarrow \text{execute}(compiled\_code, X)$
    \STATE $X_j \leftarrow \text{vars}[feature]$
    \STATE $self.name \leftarrow X_j.\text{name}$
    \STATE \textbf{Return:} $X_j$
\ELSE
    \STATE \textbf{Raise Exception}
\ENDIF
\end{algorithmic}
\end{algorithm}

\subsection*{Feature\_db}
The Feature\_db object is used to store the generated features and optimization residuals and contains the main bulk of the implementation. Algorithm \ref{alg:get_prompt} gets the prompt to pass to the LLM by sampling features according to their evaluator scores. Algorithm \ref{alg:generate_prompt} takes the selected features and combines this with the prompt template, dataset description and generated feature names and scores to construct the prompt. Algorithm \ref{alg:add_feature} adds a feature to the feature\_db. If the number of features exceeds the maximum features, it increases the generation number and resets the feature\_db. Algorithm \ref{alg:reset_feature_db} resets the feature\_db by re-initializing with the best features from the current generation. Algorithm \ref{alg:update_best_feature_set} uses XGBoost to determine whether the best feature set of the current generation is better than the best feature set from the previous generation. If so, it returns the best feature set from the current generation, otherwise it returns the best feature set from the previous generation. Finally, algorithm \ref{alg:shap_filter} uses the SHAP filter in a recursive feature elimination manner, to return the best feature set for the current generation. 

\begin{algorithm}[H]
\caption{Constructor Method of Feature\_db}
\label{alg:feat_db_constructor}
\begin{algorithmic}
\STATE \textbf{Input:}
\STATE \hspace{1em} $self$: Feature\_db object
\STATE \hspace{1em} $desc$: String dataset description
\STATE \hspace{1em} $temp$: String prompt template
\STATE \hspace{1em} $n_{\text{prompt}}$: Number of features to add to prompt
\STATE \hspace{1em} $T_0$: Temperature parameter
\STATE \hspace{1em} $K$: Temperature decay rate
\STATE \hspace{1em} $\epsilon$: Temperature decay offset
\STATE \hspace{1em} $N_{\text{max}}$: Max number of features in the feature\_db
\STATE \hspace{1em} $N$: Number of features to keep each generation
\STATE \hspace{1em} $G$: Max number of generations
\STATE \textbf{Output:}
\STATE \hspace{1em} Initialized Feature\_db object
\STATE $self.desc \leftarrow desc$
\STATE $self.temp \leftarrow temp$
\STATE $self.n\_prompt \leftarrow n_{\text{prompt}}$
\STATE $self.T\_0 \leftarrow T_0$
\STATE $self.K \leftarrow K$
\STATE $self.eps \leftarrow \epsilon$
\STATE $self.n\_max \leftarrow N_{\text{max}}$
\STATE $self.n\_keep \leftarrow N$
\STATE $self.max\_gen \leftarrow G$
\STATE $self.fts \leftarrow []$
\STATE $self.gen \leftarrow 0$
\STATE $self.residual \leftarrow []$
\STATE $self.best\_ft\_set \leftarrow []$
\STATE $self.gen\_fts \leftarrow []$
\STATE $self.X \leftarrow \text{None}$
\STATE $self.y \leftarrow \text{None}$
\end{algorithmic}
\end{algorithm}

\begin{algorithm}[H]
\caption{get\_prompt Method of Feature\_db Class}
\label{alg:get_prompt}
\begin{algorithmic}
\STATE \textbf{Input:}
\STATE \hspace{1em} $self$: Feature\_db object
\STATE \textbf{Output:}
\STATE \hspace{1em} String prompt for LLM
\STATE $n\_ft \leftarrow \text{length}(self.fts)$
\STATE $n\_prompt \leftarrow \text{min}(self.n\_prompt, n\_ft)$
\STATE $T \leftarrow self.T\_0 \cdot e^{\frac{-self.K \cdot n\_ft}{self.n\_max}} + self.eps$
\STATE $denom \leftarrow 0$
\FORALL{$f \in self.fts$}
    \STATE $score \leftarrow f.score$
    \STATE $denom \leftarrow denom + e^{\frac{score}{T}}$
\ENDFOR
\STATE $probs \leftarrow []$
\FORALL{$f \in self.fts$}
    \STATE $score \leftarrow f.score$
    \STATE $num \leftarrow e^{\frac{score}{T}}$
    \STATE $probs \leftarrow \text{insert}(probs, num / denom)$
\ENDFOR
\STATE $fts \leftarrow \text{random\_choice}(self.fts, probs, n\_prompt)$
\STATE $prompt \leftarrow \text{generate\_prompt}(fts)$
\STATE \textbf{Return:} $prompt$
\end{algorithmic}
\end{algorithm}

\begin{algorithm}[H]
\caption{generate\_prompt Method of Feature\_db Class}
\label{alg:generate_prompt}
\begin{algorithmic}
\STATE \textbf{Input:}
\STATE \hspace{1em} $self$: Feature\_db object
\STATE \hspace{1em} $fts$: Selected example features
\STATE \textbf{Output:}
\STATE \hspace{1em} String prompt for LLM
\STATE $code\_str \leftarrow ""$
\FORALL{$f \in fts$}
    \STATE $code \leftarrow f.code$
    \STATE $code\_str \leftarrow \text{add\_to\_str}(code\_str, code)$
\ENDFOR
\STATE $prompt \leftarrow \text{copy}(self.temp)$
\STATE $prompt.\text{replace}(\text{@@description@@}, self.desc)$
\STATE $prompt.\text{replace}(\text{@@examples@@}, code\_str)$
\FORALL{$f \in self.gen\_fts$}
    \STATE $score\_str \leftarrow \text{join\_as\_str}(f.name, f.score)$
    \STATE $prompt \leftarrow \text{add\_to\_str}(prompt, score\_str)$
\ENDFOR
\STATE \textbf{Return: $prompt$}
\end{algorithmic}
\end{algorithm}

\begin{algorithm}[H]
\caption{add\_feature Method of Feature\_db Class}
\label{alg:add_feature}
\begin{algorithmic}
\STATE \textbf{Input:}
\STATE \hspace{1em} $self$: Feature\_db object
\STATE \hspace{1em} $f$: Feature to add
\STATE \textbf{Output:}
\STATE \hspace{1em} Updated Feature\_db object
\STATE $self.fts.\text{insert}(f)$
\STATE $self.gen\_fts.\text{insert}(f)$
\STATE $n \leftarrow \text{length}(self.fts)$
\IF{$n \geq self.n\_max$}
    \STATE $self.gen \leftarrow self.gen + 1$
    \IF{$self.gen < self.max\_gen$}
        \STATE $self.\text{reset\_feature\_db}()$
    \ENDIF
\ENDIF
\end{algorithmic}
\end{algorithm}

\begin{algorithm}[H]
\caption{reset\_feature\_db Method of Feature\_db Class}
\label{alg:reset_feature_db}
\begin{algorithmic}
\STATE \textbf{Input:}
\STATE \hspace{1em} $self$: Feature\_db object
\STATE \textbf{Output:}
\STATE \hspace{1em} Reset Feature\_db object with best features
\STATE $best\_fts \leftarrow self.\text{update\_best\_feature\_set}()$
\STATE $self.fts \leftarrow []$
\FORALL{$f \in best\_fts$}
    \STATE $self.\text{add\_feature}(f)$
\ENDFOR
\end{algorithmic}
\end{algorithm}

\begin{algorithm}[H]
\caption{update\_best\_feature\_set Method of Feature\_db Class}
\label{alg:update_best_feature_set}
\begin{algorithmic}
\STATE \textbf{Input:}
\STATE \hspace{1em} $self$: Feature\_db object
\STATE \textbf{Output:}
\STATE \hspace{1em} List of best features
\STATE $best\_fts \leftarrow self.\text{shap\_filter}()$
\STATE $X \leftarrow \text{copy}(self.X)$
\FORALL{$f \in best\_fts$}
    \STATE $X_j \leftarrow f.\text{get\_feature}(self.X)$
    \STATE $X \leftarrow [X, X_j]$
\ENDFOR
\STATE $error \leftarrow \text{compute\_rmse}(X, self.y)$
\IF{$self.best\_ft\_set = []$}
    \STATE $self.best\_ft\_set.\text{insert}(best\_fts)$
    \STATE $self.residual.\text{insert}(error)$
\ELSE
    \IF{$error \geq self.residuals[-1]$}
        \STATE $best\_fts \leftarrow self.best\_fts[-1]$
        \STATE $error \leftarrow self.residuals[-1]$
    \ENDIF
    \STATE $self.best\_fts.\text{insert}(best\_fts)$
    \STATE $self.residuals.\text{insert}(error)$
\ENDIF
\STATE \textbf{Return: $best\_fts$}
\end{algorithmic}
\end{algorithm}

\begin{algorithm}[H]
\caption{shap\_filter Method of Feature\_db Class}
\label{alg:shap_filter}
\begin{algorithmic}
\STATE \textbf{Input:}
\STATE \hspace{1em} $self$: Feature\_db object
\STATE \textbf{Parameters:}
\STATE \hspace{1em} $\phi$: Fraction of features to prune each iteration
\STATE \textbf{Output:}
\STATE \hspace{1em} List of best features
\STATE $fts \leftarrow \text{copy}(self.fts)$
\STATE $n\_prune \leftarrow \text{max}(1, \text{floor}(\text{length}(fts) / \phi))$
\WHILE{$\text{length}(fts) > self.n\_keep$}
    \STATE $X \leftarrow \text{copy}(self.X)$
    \FORALL{$f \in fts$}
        \STATE $X_j \leftarrow f.\text{get\_feature}(X)$
        \STATE $X \leftarrow [X, X_j]$
    \ENDFOR
    \STATE $scores \leftarrow \text{compute\_shap}(X, self.y)$
    \STATE $fts \leftarrow \text{sort\_by\_scores}(fts, scores)$
    \STATE $fts \leftarrow \text{prune\_correlated}(fts)$
    \IF{$\text{length}(fts) \leq self.n\_keep$}
        \STATE \textbf{Break out of loop}
    \ENDIF
    \IF{$\text{length}(fts) - n\_prune \leq self.n\_keep$}
        \STATE $n\_prune \leftarrow \text{length}(fts) - self.n\_keep$
    \ENDIF
    \STATE $fts \leftarrow fts[:-n\_prune]$
\ENDWHILE
\STATE \textbf{Return: $fts$}
\end{algorithmic}
\end{algorithm}

\subsection*{ELATE}
The ELATE method contains the main wrapper function that interacts with the evaluators and the feature\_db. The interface follows the usual fit/transform structure. Algorithm \ref{alg:fit} populates the feature\_db using the validation feature matrix. Algorithm \ref{alg:transform} transforms a different feature matrix using the fitted feature\_db.

\begin{algorithm}[H]
\caption{Constructor Method of ELATE Class}
\label{alg:elate_constructor}
\begin{algorithmic}
\STATE \textbf{Input:}
\STATE \hspace{1em} $self$: ELATE object
\STATE \hspace{1em} $feat\_db$: Feature\_db object
\STATE \hspace{1em} $E$: List of evaluators
\STATE \hspace{1em} $M$: Large language model
\STATE \hspace{1em} $init\_code$: List of string code for initial features
\STATE \hspace{1em} $n_{\text{resp}}$: Number of responses per prompt
\STATE \textbf{Output:}
\STATE \hspace{1em} Initialized ELATE object
\STATE $self.feat\_db \leftarrow feat\_db$
\STATE $self.evaluators \leftarrow E$
\STATE $self.model \leftarrow M$
\STATE $self.seed\_fts \leftarrow init\_code$
\STATE $self.n\_resp \leftarrow n_{\text{resp}}$
\end{algorithmic}
\end{algorithm}

\begin{algorithm}[H]
\caption{fit Method of ELATE Class}
\label{alg:fit}
\begin{algorithmic}
\STATE \textbf{Input:}
\STATE \hspace{1em} $self$: ELATE object
\STATE \hspace{1em} $X$: Feature matrix dataframe to fit to
\STATE \hspace{1em} $y$: Target feature vector
\STATE \textbf{Output:}
\STATE \hspace{1em} Updated Feature\_db with fitted features
\STATE $self.feature\_db.X \leftarrow X$
\STATE $self.feature\_db.y \leftarrow y$
\FORALL{$code \in self.seed\_fts$}
    \STATE $f \leftarrow \text{Feature}(code)$
    \STATE $X_j \leftarrow f.\text{get\_feature}(X)$
    \STATE $scores \leftarrow \{\}$
    \FORALL{$E \in self.evaluators$}
        \STATE $scores[E] \leftarrow E.\text{evaluate}(X_j, y)$
    \ENDFOR
    \STATE $f.\text{add\_score}(scores)$
    \IF{any score $\in scores \neq 0$}
        \STATE $self.feat\_db.\text{add\_feature}(f)$
    \ENDIF
\ENDFOR
\STATE $gen\_no \leftarrow 0$
\WHILE{$gen\_no < self.feat\_db.max\_gen$}
    \IF{$self.feat\_db.gen\_no > gen\_no$}
        \STATE $self.model.\text{clear\_history}()$
    \ENDIF
    \STATE $gen\_no \leftarrow self.feat\_db.gen\_no$
    \STATE $prompt \leftarrow self.feat\_db.\text{get\_prompt}()$
    \STATE $resp \leftarrow self.model.\text{draw\_samples}(prompt, self.n\_resp)$
    \FORALL{$code \in resp$}
        \STATE $f \leftarrow \text{Feature}(code)$
        \STATE $X_j \leftarrow f.\text{get\_feature}(X)$
        \STATE $scores \leftarrow \{\}$
        \FORALL{$E \in self.evaluators$}
            \STATE $scores[E] \leftarrow E.\text{evaluate}(X_j, y)$
        \ENDFOR
        \STATE $f.\text{add\_score}(scores)$
        \IF{any score $\in scores \neq 0$}
            \STATE $self.feat\_db.\text{add\_feature}(f)$
        \ENDIF
    \ENDFOR
\ENDWHILE
\STATE $self.feat\_db.\text{update\_best\_feature\_set}()$
\end{algorithmic}
\end{algorithm}

\begin{algorithm}[H]
\caption{transform Method of ELATE Class}
\label{alg:transform}
\begin{algorithmic}
\STATE \textbf{Input:}
\STATE \hspace{1em} $self$: ELATE object
\STATE \hspace{1em} $X$: Feature matrix dataframe to transform
\STATE \textbf{Output:} 
\STATE \hspace{1em} Transformed feature matrix dataframe
\STATE $fts \leftarrow self.feat\_db.best\_fts[-1]$
\STATE $X' \leftarrow \text{copy}(X)$
\FORALL{$f \in fts$}
    \STATE $X_j \leftarrow f.\text{get\_feature}(X)$
    \STATE $X' \leftarrow [X', X_j]$
\ENDFOR
\STATE \textbf{Return:} $X'$
\end{algorithmic}
\end{algorithm}

\newpage
\section{Evaluator Details}\label{app:eval}

\subsection*{Granger Causality}
Granger Causality is a statistical measure which can be used to evaluate whether one time-series has predictive power over another. If past values of a time-series $X$, provide information that helps to predict future values of the target $Y$; and $X$ has unique information about the future of $Y$, then $X$ is said to Granger-cause $Y$. We define an Vector Autoregression (VAR) linear model, containing both lagged values of $Y$ and $X$:
\begin{gather}\label{eq:var}
    Y_t = \sum_{i=1}^pa_i Y_{t-i} + \sum_{i=1}^pb_i X_{t-1} + \epsilon
\end{gather}
\noindent Where $p$ is the number of lagged observations included in the model. $X$ has predictive power of $Y$ if the predictive performance is increased by including the $X$ terms, or some of the $X$ terms plays a significant role in the prediction. Based on this, we have the following hypothesis:
\begin{gather*}
    H_0: b_1 = b_2 = \dots = b_p = 0 \\
    H_1: \text{at least one } b_1, b_2, \dots, b_p \neq 0
\end{gather*}

\noindent If we denote $RSS^{H_1}$, as the residual sum of squares (RSS) of \eqref{eq:var}; and $RSS^{H_0}$ as the RSS of a restricted model excluding $X$: $Y_t = \sum_{i=1}^pa_i Y_{t-i} + \epsilon$, then the F-statistic is:
$$
F = \frac{(RSS^{H_0} - RSS^{H_1}) / p}{RSS^{H_1} / (n - p)}
$$
Note that $(RSS^{H_0} - RSS^{H_1})$ measures the reduction in error when $X$ is included in the model. The $p$-value of the F-statistic tells us whether the observed improvement is likely to be due to chance. Typically, when $p < 0.05$, the improvement is deemed unlikely to be a result of chance and therefore we can say that $X$ Granger causes $Y$. Otherwise, there is no evidence that $X$ Granger causes $Y$. Hence, if $p \geq 0.05$, then we set $E_{GC}(X) = 0$, otherwise we set $E_{GC}(X) = |b_1|$, which indicates the magnitude of the strength of the causality. To compute granger causality we use the test\_causality method within the statsmodels package.

\subsection*{Mutual Information}
Mutual information tells us the amount of information that can be learned about one random variable by observing another. In the context of time-series, we can think of it as quantifying the amount of information, we can learn about the target $Y$ by observing a feature $X$. Unlike Granger causality, mutual information is capable of capturing non-linear dependencies between the two time-series. The mutual information of two random variables is:
$$
I(X,Y) = \int_Y \int_X P_{X,Y}(x,y) \log\left(\frac{P_{X,Y}(x,y)}{P_X (x) P_Y (y)}\right)dx dy
$$
\noindent Where $P_{X,Y}$ is the joint probability density function of $X$ and $Y$ and $P_X$ and $P_Y$ are the marginal probability density functions of $X$ and $Y$. Higher values of $I(X,Y)$ correspond to a greater dependency between $X$ and $Y$, while $I(X,Y) = 0$ corresponds to the case when $X$ and $Y$ are independent. This is easy to see: if $X$ and $Y$ are independent, $P_{X,Y} = P_X(x) P_Y(y)$ and therefore $\log 1 = 0$. To compute mutual information, we use mutual\_info\_regression from scikit-learn.


\end{document}